\UseRawInputEncoding
\pdfoutput=1
\documentclass[review]{elsarticle}

\usepackage{hyperref}
\usepackage{bm}
\usepackage{array}
\usepackage[export]{adjustbox}
\usepackage{amsmath}  
\usepackage{amssymb}  
\usepackage{booktabs} 
\usepackage{float}      
\usepackage{multirow}   
\usepackage{mathrsfs}   
\usepackage{subfigure}  
\usepackage{indentfirst}

\usepackage{subfigure}
\usepackage{graphicx}					
\graphicspath{{./picture/}}
\DeclareGraphicsExtensions{.jpg}

\journal{Journal of Mechatronics}

\begin{document}

\begin{frontmatter}

\title{Flying Trot Control Method for Quadruped Robot Based on Trajectory Planning}





\author[SDUAddress]{Hongge Wang }
\ead{wanghg1992@outlook.com}
\author[SDUAddress]{Hui Chai \corref{cor1} }
\ead{chaimax@sdu.edu.cn}
\author[YOBOTICSAdress]{Bin Chen }
\ead{15820000319@139.com}
\author[SDUAddress,YOBOTICSAdress]{Aizhen Xie }
\ead{xieaizhensdu@163.com}
\author[SDUAddress]{Rui Song }
\ead{rsong@sdu.edu.cn}
\author[BeifangAdress]{Bo Su }
\ead{bosu@noveri.com.cn}


\cortext[cor1]{Corresponding author}


\address[SDUAddress]{Robotics Research Center, School of Control Science and Engineering,Shandong University, Shandong Province, China }

\address[YOBOTICSAdress]{Shandong Youbaote Intelligent Robot Co., Ltd, Jinan, Shandong Province, China}
\address[BeifangAdress]{China North Vehicle Research Institute, Beijing, China}

%

\begin{abstract}
An intuitive control method for the flying trot, which combines offline trajectory planning with real-time balance control, is presented. The motion features of running animals in the vertical direction were analysed using the spring-load-inverted-pendulum (SLIP) model, and the foot trajectory of the robot was planned, so the robot could run similar to an animal capable of vertical flight, according to the given height and speed of the trunk. To improve the robustness of running, a posture control method based on a foot acceleration adjustment is proposed. A novel kinematic based CoM observation method and CoM regulation method is present to enhance the stability of locomotion. To reduce the impact force when the robot interacts with the environment, the virtual model control method is used in the control of the foot trajectory to achieve active compliance. By selecting the proper parameters for the virtual model, the oscillation motion of the virtual model and the planning motion of the support foot are synchronized to avoid the large disturbance caused by the oscillation motion of the virtual model in relation to the robot motion. The simulation and experiment using the quadruped robot Billy are reported. In the experiment, the maximum speed of the robot could reach 4.73 times the body length per second, which verified the feasibility of the control method.\end{abstract}


\end{frontmatter}

\section{Introduction}
\indent The legged robot is more adaptable to the non-structural environment than the wheeled robot and the tracked robot. The quadruped robot has better stability than the biped robot and has a simpler structural complexity than the hexapod robot, so the quadruped robots have attracted increasingly more attention from scholars. \\	

\indent Improving the speed of movement is a trend in quadruped robots. The common gaits of quadruped robots include crawl, trot, pace, bound, and gallop. Due to the good symmetry, small posture fluctuations, and the ability to balance the movement speed and energy consumption, the trot gait is widely used in the movement of the quadruped robot. 


\indent The flying trot gait is a trot gait with flying phases that can achieve higher moving speeds. The quadruped robots that run with a flying trot gait are Quadruped, designed by the Massachusetts Institute of Technology \cite{Legged_robots_that_balance, Quadruped_Hopping_in_legged_systems}, Kolt \cite{Kolt_System_design, Kolt_Intelligent_control, Kolt_Thrust_control} of Stanford University and Ohio State University, HyQ \cite{HyQ_Is_active_impedance} of the Italian Institute of Technology, StarlETH \cite{StarlETH_Co_design_and_control, StarlETH_Control_of_dynamic_gaits} of the Zurich Federal Institute of Technology, Cheetah cub \cite{Cheetah_cub_Towards_dynamic_trot} of Lausanne Federal Institute of Technology, Boston Dynamics' BigDog \cite{BigDog, Bigdog_the_rough_terrain}, LS3 \cite{Meet_Boston_dynamics_LS3} and Spot \cite{Introducing_Spot}; in addition, Billy \cite{Billy_quadruped_robot} is the quadruped robot used in the experiments of this paper, as shown in Fig. 1a, which is cooperatively developed by Shandong University and Shandong Youbaote Intelligent Robot Co., Ltd. The control methods of the flying trot gait that is used are summarized as follows.

%


\begin{figure}[htbp] 
  \centering 
  \subfigure[Billy quadruped robot]{ 
    \includegraphics[width=1.8in]{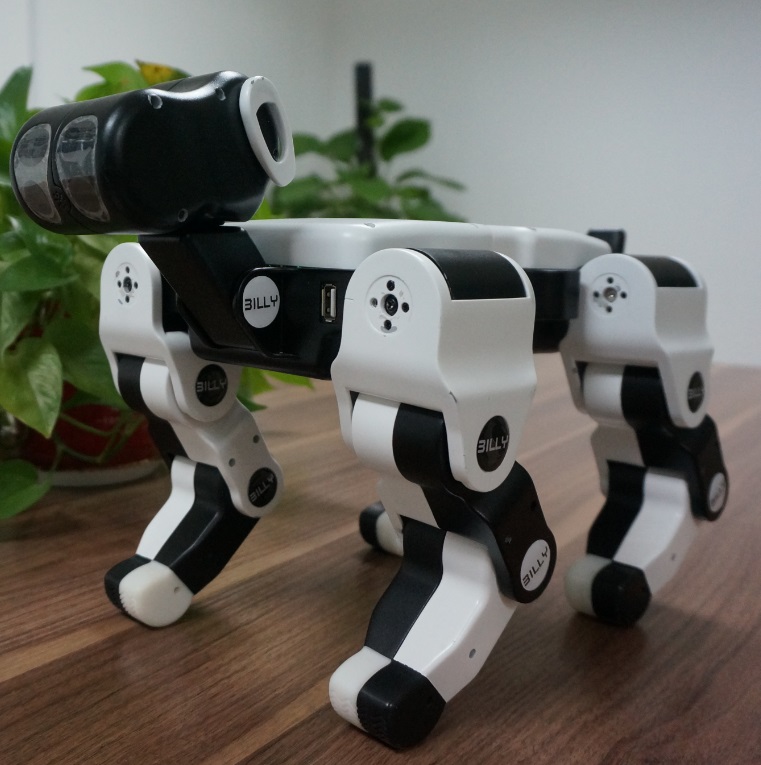} 
  } 
  \subfigure[Simulation model]{ 
    \includegraphics[width=1.85in]{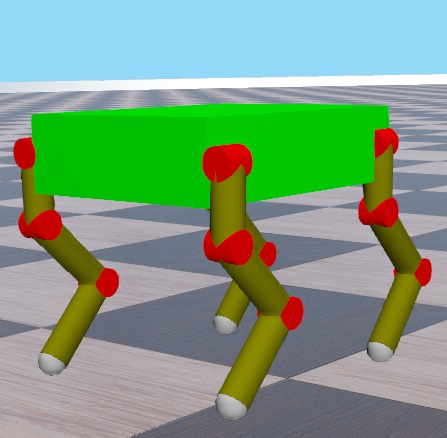} 
  } 
  \caption{Billy, the quadruped robot used in the experiments, and its simulation model in Webots} 
\end{figure}


\indent The researchers of Quadruped put forward the control method, which divided the control of the robot into three parts: trunk height control, posture control and speed control, and realized a stable trot movement. Controlling the energy of the robot in a vertical direction realized the flying of the Quadruped robot in the trot movement \cite{Legged_robots_that_balance, Quadruped_Trotting_pacing_and, Quadruped_Running_on_four_legs}. However, because the control method of vertical flying is closely combined with the pneumatic control method of the legs, it is not universal.

\indent By controlling the touch position of the swing foot to control the robot's forward velocity, lateral velocity, heading angular velocity, in addition to controlling the energy injected to the support leg spring to control the trunk height and controlling the joints' torque of the support leg to control the pitch angle and roll angle, Kolt can move stably with a trot gait [8]. When the descent speed of the support foot reaches a certain value, Kolt uplifts the support foot to achieve flying in the trot gait [9]. However, the researchers did not detail the specific process of the descent control and  uplift control of the support foot, nor did they establish the specific model of the flying control.


\indent The motion control of the quadruped robot StarlETH is divided into trunk control, swing foot control and support foot control. The swing foot control is mainly used to select the proper touch position to control the robot speed and make the robot recover from an external disturbance. The trunk control mainly calculates a desired virtual six-dimensional force acting on the trunk from the deviation between the control target and the actual situation of trunk and then calculates a group of optimal forces acting on the support feet through an optimization method. The support foot control aims to calculate the corresponding joint torque from the desired optimal foot forces of the support legs and execute the torque. The flying phase is achieved by controlling the height of the trunk in the trot gait \cite{StarlETH_Control_of_dynamic_gaits}. However, the researchers of StarlETH did not details the process of the height control of the trunk to achieve the flying phase.

\indent The quadruped robot Cheetah cub uses open-loop control. Its researchers control the movement of the quadruped robot with the method of a central pattern generator (CPG) based on the joint workspace. Then, through a large number of experiments, the parameters of CPG are optimized to improve the forward speed of the quadruped robot \cite{Cheetah_cub_Towards_dynamic_trot}. However, the optimal parameters of different quadruped robots are different, and the process of parameter optimization is complex. Moreover, since the quadruped robot uses the CPG method based on joint space, this method may not be applicable once the joint configuration of the robot is changed. 

\indent The flying trot of the quadruped robot HyQ is based on active compliant control. The desired trajectory of the foot is planned by the CPG method based on the foot workspace \cite{HyQ_A_reactive_controller_framework}, and the desired joint angle is obtained by inverse kinematics. PD control is then used in joint space to make the actual joint angle follow the desired joint angle and achieve active compliance. They stabilized the flying trot by choosing an appropriate step frequency and duty factor \cite{Scout_Some_finite_state_aspects} to make the motion of the quadruped robot synchronize with the oscillation of the virtual spring-mass system in the support phase \cite{HyQ_Is_active_impedance}. However, the HyQ researchers did not provide the method for selecting the step frequency and duty factor.

The flying trot control methods of BigDog, LS3, and Spot have not been publicly disclosed. The current information of a quadruped robot that can run with a flying trot gait is shown in table 1. Cheetah cub has a ratio of speed to body length greater than Billy, but it has no the rolling rotary joint, so it cannot move flexibly in all directions, and it is lighter because its power supply is external.


\begin{table}[htbp]
\centering
\caption{Flying trot speed of quadruped robots}
\begin{tabular}{lllll}
	\toprule
	Year & Robot 		& Mass (kg) & BL (m)  & Speed (BL/s) \\
	\midrule                           
	1990 & Quadruped 	& 38  & 0.78 	& 2.82 \\
	2004 & Kolt 		& 80  & 1.75 	& 0.63 \\
	2010 & HyQ 			& 70  & 1.3 	& 1.3 \\
	2013 & StarlETH 	& 23  & 0.5	& * \\
	2013 & Cheetah cub 	& 1.1  & 0.205	& 6.9 \\
	     &  BigDog  	& 109  & 1.1	& 1.82 \\
	2004-2015 &  LS3  	& *    & *		& * \\
	     &  Spot  		& 72  & *		& * \\
	2016 & Billy 		& 1.9  & 0.167	& 4.73 \\
	\bottomrule
\end{tabular}
\end{table}


\indent This paper takes quadruped as the research object, analyses the movement process of a quadruped in the running of the flying trot gait with the SLIP model, and puts forward the gait control method of the flying trot based on the position planning of the foot workspace, including the control of the robot's eject, landing buffer and forward speed. To increase the robustness of movement, a new posture control method based on position was proposed, and the method of virtual model control was used to carry out active compliance control to reduce the impact force when the robot interacts with the environment.

The advantages of the method proposed in this paper are as follows. (1) The implementation method of the flying trot gait based on position planning was proposed, and the touchdown time was predicted in planning to make buffer actions in advance and reduce the impact of landing. (2) A novel position-based posture control method was proposed, which can be used to enhance the balance of the robot. (3) To further reduce the impact of landing in running, the method of the leg virtual model was used for active compliance control, and a reasonable selection method of kp and kd for virtual spring and damping was proposed for the running gait.

\indent The simulations in Webots and the experiments of Billy prove that the method proposed in this paper is effective and feasible. Fig. 1b is the simulation model of Billy in Webots.
 
The rest of this article is summarized as follows. In the second chapter, a simplified structure model of Billy is introduced. In the third chapter, the foot trajectory planning method of the flying trot gait is described in detail. The fourth chapter describes the posture control method based on the position control, the control method of active compliance and the selection method of the related parameters. In chapter 5, the results of the simulations and experiments are given. The sixth chapter presents the summary of the control method in this article.

%
%

%
%


\section{Quadruped Model}

\begin{figure}[htbp]
\centering
\includegraphics[width=3.2in]{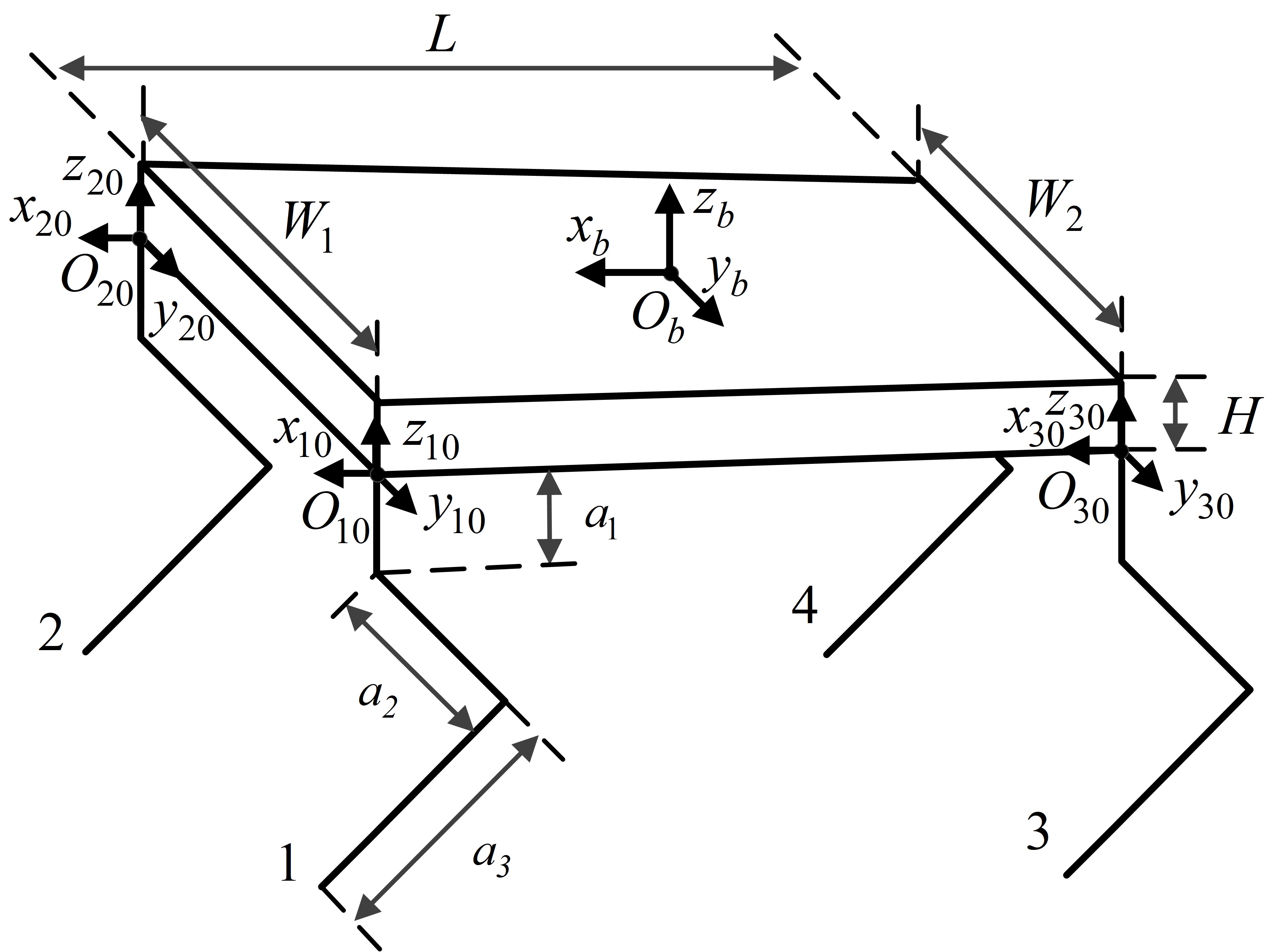}
\caption{Structure of simplified model of Billy}
\end{figure}

\indent The simplified structure model of the quadruped robot is shown in Fig. 2. The robot consists of a trunk and four legs. Each leg contains a roll joint and two pitch joints. The left fore leg (LF), right fore leg (RF), left hind leg (LH) and right hind leg (RH) are numbered 1, 2, 3 and 4, respectively. In the trot gait, the diagonal legs of the robot always move in sync. To simplify the description in the following chapters, the four legs are divided into two groups according to the diagonal, and the left fore leg and right hind leg are group L, while the right fore leg and left hind leg are group R. Write the support leg group as S and swing leg group as W. The size of the robot's trunk and the length of each link are shown in table 2.

A coordinate system of $O_b$ is established at the geometric centre of the robot's trunk. The direction of the $x$-axis points to the forward direction, the z-axis points to the upward direction, and the direction of y-axis is obtained by using the right-hand rule with the z-axis and x-axis. Set up the coordinate system $O_{i0}$, where the robot's legs and trunk are connected. The direction of the coordinate axis is the same as that of the trunk coordinate system.

\begin{table}[htbp]
\centering
\caption{The size of the robot's trunk and the length of each link}
\begin{tabular}{lll}
	\toprule
	Variables & Description & Value (m) \\
	\midrule
	$L$ & Body length 	 	& 0.167	\\	
	$W_1$ & Shoulder width 	& 0.162	\\	
	$W_2$ & Hip width 	 	& 0.142	\\	
	$a_1$ & Hip link length & 0.046	\\	
	$a_2$ & Thigh length 	& 0.066	\\	
	$a_3$ & Crus length  	& 0.065	\\	
	\bottomrule
\end{tabular}
\end{table}

\section{Offline Planning}
\indent In this section, the SLIP model is used to analyse the dynamic characteristics in the vertical direction of a quadruped running, and the trajectory of the centre of mass (CoM) of a quadruped running robot is planned based on these dynamic characteristics. Based on robot's CoM trajectory and some parameters related to dynamic characteristics, the foot trajectory planning method of the flying trot gait is introduced in detail. At the planning level, the landing prediction of the running robot is made, and the landing buffer is prepared in advance. Then, the landing buffer and the ejection from the ground are completed to realize the control of the quadruped robot flying. In the end, the method of controlling the speed of the trunk by planning and selecting the horizontal landing point and the method of controlling the heading angular velocity of the robot by adjusting the trajectory of the foot end by using the rotation matrix are given.

\subsection{SLIP Model of the Running Animals}
The SLIP model has been an effective and accurate model to describe all kinds of animals running for a long time. In the SLIP model, the animal's support foot is equivalent to an elastic link, and the animal's trunk is equivalent to a mass point on the top of the elastic link. The mass point moves by the interaction between the elastic link and the ground. The running of an animal in one step can be divided into four processes: landing buffering (A), ejection from the ground (B), ascending (C) and descending (D), as shown in Fig. 3. SP stands for support phase, and WP stands for swing phase.

\begin{figure}[htbp]
\centering
\includegraphics[width=2.5in]{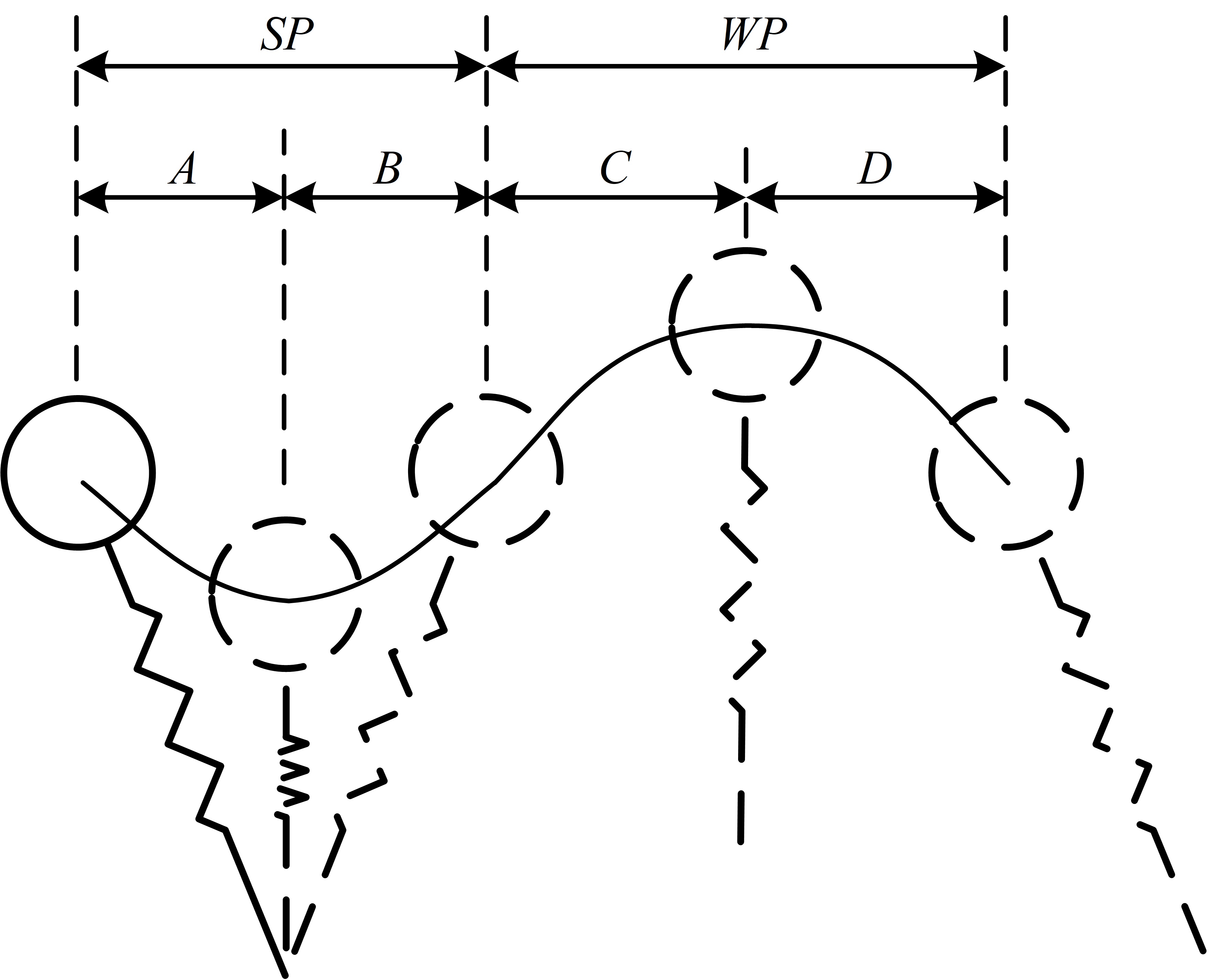}
\caption{SLIP model of the running animals}
\end{figure}

In phase A, the animal finishes landing and shrinks its legs to cushion the impact between the foot and ground. In phase B, its legs are lengthened gradually to accelerate its trunk upward. In stage C, the legs are constrained to a desired length, and the upward momentum of the trunk will lift the animal completely off the ground and up to the highest position. In stage D, the trunk begins to descend, and the animal eventually lands.

\subsection{CoM Trajectory Planning}
Inspired by the running movement of animals, the CoM trajectory of the robot body can be designed to achieve the landing buffering and ejection from the ground. To simulate the running of animals, the CoM trajectory of the robot in a flying trot gait cycle was designed, as shown by the black solid line in Fig. 4. There are two flying phases (FP) in the figure because in a flying trot gait cycle, the legs of group L and group R completed a landing buffering and an ejection, respectively. SP1 represents the support stage of the legs of group L, while TP1 represents the swinging stage of the legs of group L. SP2 and TP2 represent the corresponding phases of the legs of group R.   

\begin{figure}[htbp]
\centering
\includegraphics[width=3.5in]{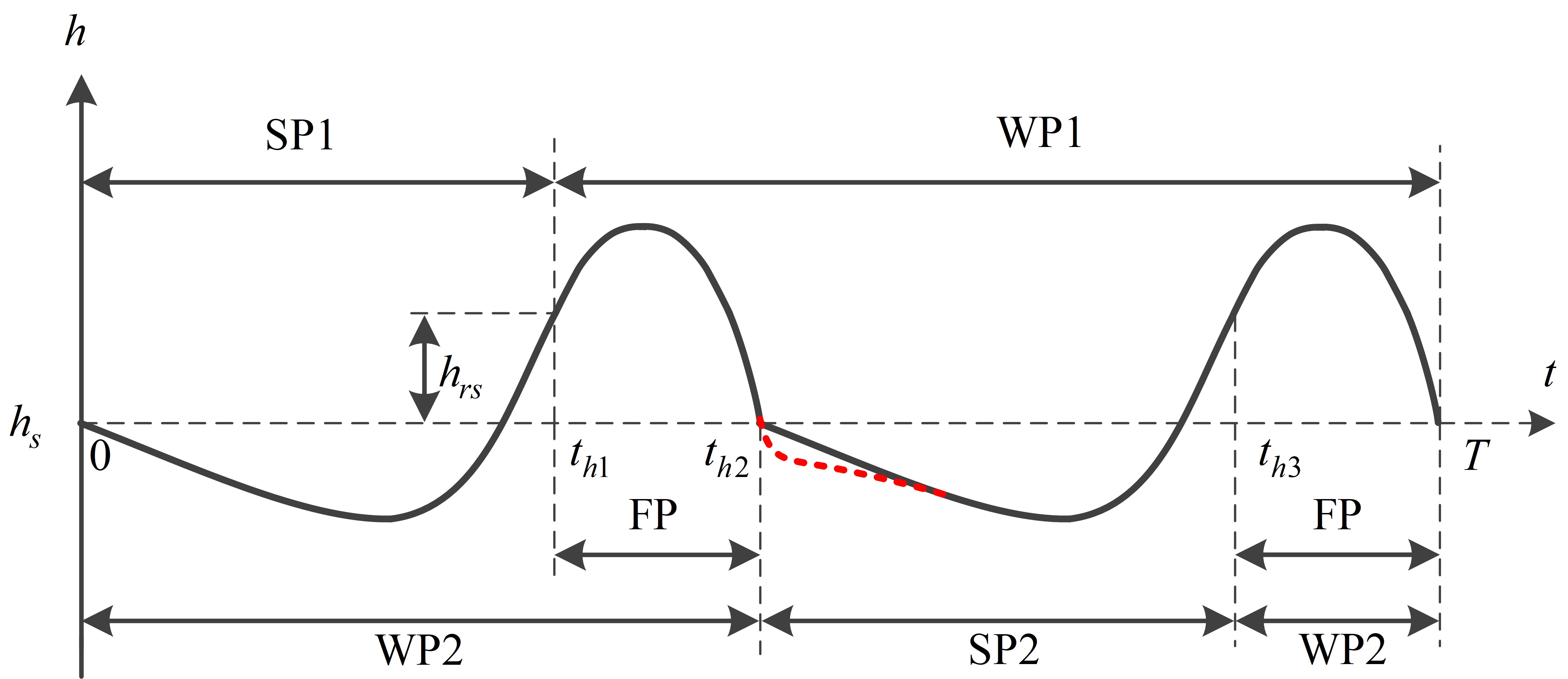}
\caption{CoM trajectory in a gait cycle. The planning speed of the CoM may be discontinuous at $t_{h2}$, shown as the black solid line, because the foot trajectory planning may not cushion the landing impact completely due to the performance limit of actuators. However, the actual speed of the CoM will be continuous, shown as the red dashed line, by the using of passive compliant mechanism and active compliance control for legs.}
\end{figure}


\subsection{Foot Trajectory Planning in the z-axis}
Because of the symmetry of the trot gait, the robot has good stability in the trot gait movement. Assuming that the trunk is horizontal and that there is no slip between the support feet of the robot and the ground during the movement, the trajectory of support feet relative to the CoM of the trunk can be planned according to the CoM trajectory of the trunk. The swing feet trajectory can be obtained from the robot's control target. In this paper, all the foot trajectory planning is based on the coordinate system $O_{i0}$, $i=1,2,3,4$. Taking the LF leg as an example, the foot trajectory planning on the z-axis is shown in Fig. 5. The LF foot lands at time 0 and begins to lift up to cushion the landing impact. It then moves down a distance of $h_{sd}$ to accelerate the trunk upward. During $t_{z1} \sim t_{z2} $, it accelerates the trunk upwards to get the robot off the ground. During $t_{z2} \sim t_{z3}$, it moves up to $h_{wa}$ to avoid obstacles on the ground. During the period of $t_{z3} \sim T$, the foot is moved downward to prepare for landing and then makes a buffer action in advance since landing can be anticipated near the time of T. At time $T$, the gait cycle ends. To simplify the description in following chapter, we define $\Delta t_{zi}=t_{zi}-t_{z(i-1)}$, $i=1,2,3,4$.\\

\begin{figure}[htbp]
\centering
\includegraphics[width=3.5in]{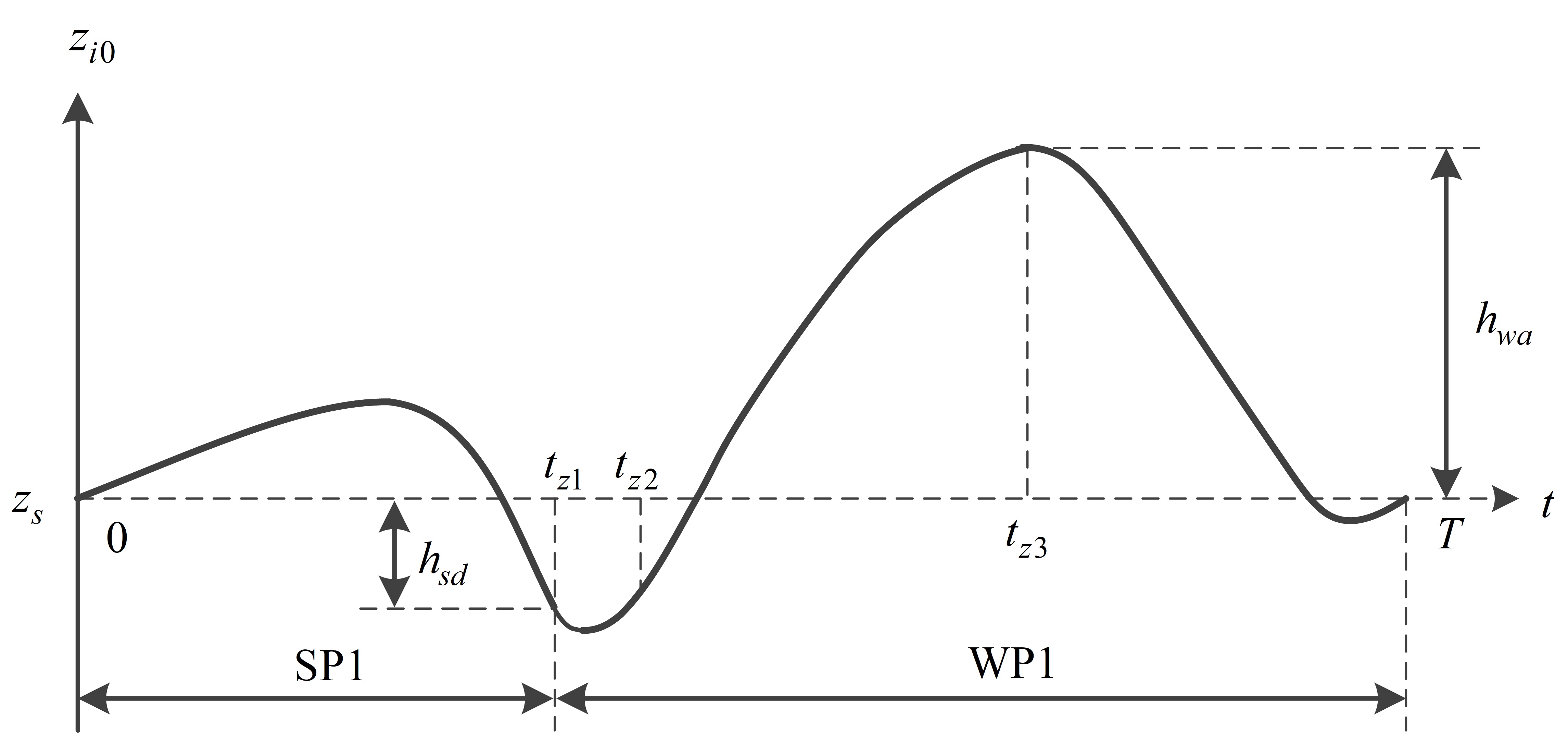}
\caption{Foot trajectory on the z-axis in a gait cycle}
\end{figure}


\indent Based on a large number of experiments and analysis, we find that the foot position and speed in the z-axis at $t=0,t_{z1},t_{z2},t_{z3},T$ and the interval time $\Delta t_{z1},\Delta t_{z2},\Delta t_{z3},\Delta t_{z4}$ between these points have a great influence on the movement of the robot. To determine these variables that shape the foot trajectory, we propose some control variables, which are shown in table 3. The values of control variables are determined by the control targets.

\begin{table}[htbp]
\centering
\caption{Control variables of the foot trajectory}
\begin{tabular}{ll}
	\toprule
	Control Variables & Description \\
	\midrule
	$f$ & Step frequency 	 \\
	$v_x$ & Forward speed 		 \\
	$z_{s}$ & Foot position in z-axis when robot lands 	 \\
	$h_{wa}$ & Ascending height of foot in the swing phase			 \\
	$h_{sd}$ & Descending height of foot in the support phase 	 \\
	$c_{1}$  &  The ratio of $v_{x1}$ to $v_{z1}$  \\
	$c_{2}$ &  The ratio of $v_{z0}$ to $v_{h4}^{-}$   \\
	$c_{3}$  &  The ratio of the contraction acceleration to g   \\
	$c_{4}$ & The ratio of $v_{z2}$ to $v_{z1}$ 		 \\
	$c_{5}$ & The ratio of $\Delta t_{z3}$ to $\Delta t_{z3}+\Delta t_{z4}$ 		 \\
	\bottomrule
\end{tabular}
\end{table}


\subsubsection{step frequency}
In this paper, the step period $T$, the reciprocal of the step frequency $f$, is defined as the time during which each leg of a quadruped robot experiences one support phase and one swinging phase \cite{Scout_Some_finite_state_aspects}.

\begin{equation}
 2t_{FP}+2\Delta t_{z1}= \frac{1}{f}
\end{equation}

When the step frequency is higher, the robot will be more stable, but it will also require a higher dynamic performance of the joints of the robot. When selecting the walking frequency of the robot, the walking frequency of the animal can be referred to. It has been found that the frequency with which animals usually trot depends on the mass of the animal itself \cite{Speed_stride_frequency_and}:
\begin{equation*}
 f_{prefered}=3.35m^{-0.13}
\end{equation*}

\subsubsection{forward speed}
We want the robot to move at a constant speed in the forward direction, and denote it as $v_x$. 

Similar to the selection of step frequency, when selecting the forward speed for the robot, the commonly used forward speed of animals with the same mass can also be referred to as \cite{Speed_stride_frequency_and}
\begin{equation*}
 v_{prefered}=1.09m^{0.222}
\end{equation*}

\subsubsection{foot position on the z-axis when the robot lands}
This variable represents the distance between the CoM of the trunk and the support foot on the z-axis when the robot lands on the ground. In this paper, the flying trot's standing height is defined as $-z_{s}$. The standing height should be selected according to the foot workspace of the robot to ensure that the trajectory planning of robot's foot will not exceed the workspace of the foot.

The standing height has a great influence on the maximum rotation speed and the maximum torque of the leg joints in the movement: the higher the standing height is, the higher the maximum rotation speed of the leg joints and the smaller the maximum joint torque of the support leg that are required. Therefore, it is required to choose the standing height appropriately to balance the maximum joint rotation speed and maximum joint torque.

\subsubsection{ascending height of the foot in the swing phase}
In the real world, there are often many obstacles on the ground. It is easy to trip up a robot when the robot's swinging foot meets the obstacle. To reduce the possibility of meeting obstacles in the movement of the swinging foot, the step height of the swinging foot should be set as a larger value. However, if the step height is too large, the average foot speed of the robot on the z-axis will be very large, which may cause a large impact on the ground when landing and weaken the stability of the robot. Therefore, the step height should be reasonably selected according to the actual situation of the terrain and the robot size.

\subsubsection{descending height of the foot in the support phase}
In the support phase of the robot, the support foot should accelerate downward to accelerate the robot trunk upward and achieve the ejection from the ground. This variable represents the difference between the value of $p_{z0}=z_{s}$ when the robot lands and the value $p_{z1}$ when robot ejects from the ground. The larger this variable is, the more the robot will extend its support leg when ejecting from the ground.

\subsubsection{the ratio of \texorpdfstring{$v_{x1}$}{pdfstring} to \texorpdfstring{$v_{z1}$}{pdfstring}}

When the robot ejects from the ground, the angle between the speed vector of the robot and the horizontal ground is expected to be within an appropriate range. If the angle is too large, running will consume too much energy due to the large fluctuation of the trunk moving up and down. If the angle is too small, the flight time will be too short to provide sufficient time for the leg to swing. Therefore, we need to set the ratio of $v_{x1}$ to $v_{z1}$, represented by $c_{1}$, to an appropriate value. We obtain $v_{h1}=-v_{z1}=-v_{x1}/c_{1}=v_x/c_{1}$. The CoM trajectories of the robot trunk in plane ${{xoz}}$ with different $c_1$ are shown in Fig. 6.\\
\indent Since the robot is in free fall during $t_{h1} \sim t_{h2}$, ${v_{h2}^{-}}^{2}-{v_{h1}}^{2}=-2gh_{sd}$ and $t_{FP}=\frac{v_{h2}^{-}-v_{h1}}{g}$ is derived. Then , $t_{FP}$ and $v_{h2}^{-}$ can be calculated by\\

\begin{equation}
 v_{h2}^{-}= \sqrt{ \frac{v_x^{2}}{{c_{1}}^{2}}-2gh_{sd}} 
\end{equation}

\begin{equation}
 t_{FP}= -\frac{v_x}{c_{1}g}+\sqrt{ \frac{v_x^{2}}{{c_{1}}^{2}g^{2}}-\frac{2h_{sd}}{g}} 
\end{equation}

\begin{figure}[htbp]
\centering
\includegraphics[width=3.5in]{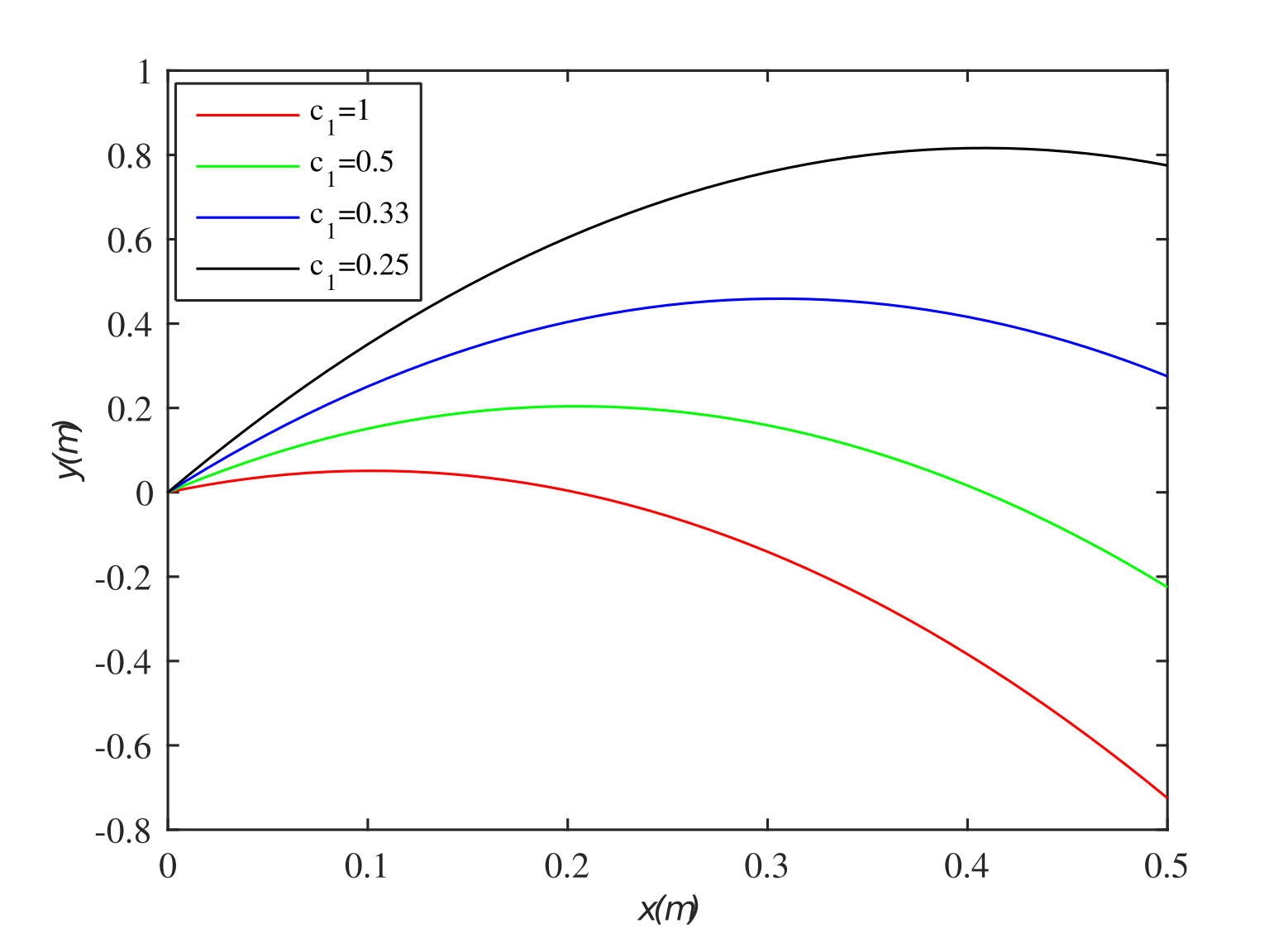}
\caption{The CoM trajectories of the robot trunk in plane ${{xoz}}$ with different $c_1$}
\end{figure}

\subsubsection{the ratio of \texorpdfstring{$v_{z0}$}{pdfstring} to \texorpdfstring{$v_{h4}^{-}$}{pdfstring}}

\indent To reduce the impact of landing, the robot should lift its foot up when landing. If $c_{2}\in [-1,0]$, the ratio of $v_{z0}$ to $v_{h4}^{-}$ is set to -1, the speed of lifting the foot equals the speed of the trunk CoM falling, and the impact is completely absorbed by the planning. If $c_{2}$ is set to 0, it means that the trajectory planning will not help in absorbing the impact. Usually, we set $c_{2}$ to a number between $-1$ and $0$, and the impact part that is not absorbed by planning can be weakened by the active compliance control or the passive compliance of the mechanical structure. Combining equation (2), $v_{h2}^{-}=v_{h4}^{-}$ and $v_{z0} = c_{2} v_{h4}^{-}$, we obtain

\begin{equation}
 v_{z0}=c_{2} \sqrt{ \frac{v_x^{2}}{{c_{1}}^{2}}-2gh_{sd}} 
\end{equation}

\subsubsection{the ratio of the shrinking acceleration to g and the ratio of \texorpdfstring{$v_{z2}$}{pdfstring} to \texorpdfstring{$v_{z1}$}{pdfstring}}
When ejecting, the foot must lift with an acceleration $c_{3}g$, greater than the acceleration of gravity $g$, to go to a flying phase, $c_{3} \in (-\infty,-1]$. The process of leg shrinking with an acceleration of $c_{3}g$ ends when the upward speed of the foot goes up to $v_{z2}=c_{4}v_{z1}$. Because $v_{z1}$ is negative, we limit $c_{4}$ in $(-\infty,0]$ to make sure that the foot leaves the ground in time. Since $p_{z1}$, $v_{z1}$ was given before, we can obtain 

\begin{equation}
 \Delta t_{z2}=\frac{(1-c_4)v_x}{c_1c_3g}
\end{equation}

\begin{equation}
p_{z2}=z_{s}-h_{sd}+\frac{({c_4}^2-1)v_x^2}{2c_3g{c_1}^2}
\end{equation}


\subsubsection{the ratio of \texorpdfstring{$\Delta t_{z3}$}{pdfstring} to \texorpdfstring{$\Delta t_{z3}+\Delta t_{z4}$}{pdfstring}}
In the swing phase, the foot ascends to the desired position $z_{s}+h_r$ for the desired step height and then descends to $z_{s}$ to prepare for landing. The time of ascending $\Delta t_{z3}$ is usually selected to be equal to the time of dropping $\Delta t_{z4}$ by setting $c_5=0.5$, $c_5\in (0,1)$. Combining equation (1) and equation (3), we obtain 

\begin{equation}
\Delta t_{z1}=\frac{1}{2f}+\frac{v_x}{c_1g}-\sqrt{\frac{v_x^2}{{c_1}^2g^2}-\frac{2h_{sd}}{g}}
\end{equation}

\begin{table}[htbp]
\centering
\caption{Variables of foot trajectory}
\begin{tabular}{ll}
	\toprule
	Variables & Computational Formula \\
	\midrule
	$\Delta t_{z1}$ & $\frac{1}{2f}+\frac{v_x}{c_1g}-\sqrt{\frac{v_x^2}{{c_1}^2g^2}-\frac{2h_{sd}}{g}}$ 	\\
	$\Delta t_{z2}$ & $\frac{(1-c_4)v_x}{c_1c_3g}$ 			 \\
	$\Delta t_{z3}$ & $c_5\left(\frac{1}{2f}-\frac{v_x}{c_1g}+\sqrt{\frac{v_x^2}{{c_1}^2g^2}-\frac{2h_{sd}}{g}}-\frac{(1-c_4)v_x}{c_1c_3g}\right)$ 	 \\
	$\Delta t_{z4}$ & $\left(1-c_5\right)\left(\frac{1}{2f}-\frac{v_x}{c_1g}+\sqrt{\frac{v_x^2}{{c_1}^2g^2}-\frac{2h_{sd}}{g}}-\frac{(1-c_4)v_x}{c_1c_3g}\right)$  \\
	$p_{z0}=p_{z4}$ &  $z_{s}$ \\
	$p_{z1}$  &  $z_{s}-h_{sd}$   \\
	$p_{z2}$ & $z_{s}-h_{sd}+\frac{({c_4}^2-1)v_x^2}{2c_3g{c_1}^2}$ 		 \\
	$p_{z3}$ & $z_{s}+h_{wa}$ 		 \\
	$v_{z0}=v_{z4}$ &  $c_{2} \sqrt{ \frac{v_x^{2}}{{c_{1}}^{2}}-2gh_{sd}}$ \\
	$v_{z1}$  &  $v_x/c_{1}$   \\
	$v_{z2}$ & $c_{4}v_x/c_{1}$ 		 \\
	$v_{z3}$ & $0$ 		 \\
	\bottomrule
\end{tabular}
\end{table}

Combining equation 5, equation 7, $c_5=\frac{\Delta t_{z3}}{\Delta t_{z3}+\Delta t_{z4}}$ and $\Delta t_{z3}+\Delta t_{z4}=\frac{1}{f}-\Delta t_{z1}-\Delta t_{z2}$, we obtain

\begin{equation}
\Delta t_{z3}=c_5\left(\frac{1}{2f}-\frac{v_x}{c_1g}+\sqrt{\frac{v_x^2}{{c_1}^2g^2}-\frac{2h_{sd}}{g}}-\frac{(1-c_4)v_x}{c_1c_3g}\right)
\end{equation}

\begin{equation}
\Delta t_{z4}=\left(1-c_5\right)\left(\frac{1}{2f}-\frac{v_x}{c_1g}+\sqrt{\frac{v_x^2}{{c_1}^2g^2}-\frac{2h_{sd}}{g}}-\frac{(1-c_4)v_x}{c_1c_3g}\right)
\end{equation}


\indent After the above analysis, all the position and speed information of selected points on the foot trajectory and all the interval times between these points can be calculated by the control variables, as shown in table 4.




\subsection{Foot Trajectory Planning on the x-axis and y-axis}
The control of the moving speed of the robot body is a main target of robot control, which is mainly determined by the planning of the robot foot on the $x$-axis and $y$-axis. The foot trajectory planning of the robot on the $x$-axis is similar to that on the $y$-axis. Take the $x$-axis as an example.

\begin{figure}[htbp]
\centering
\includegraphics[width=3.5in]{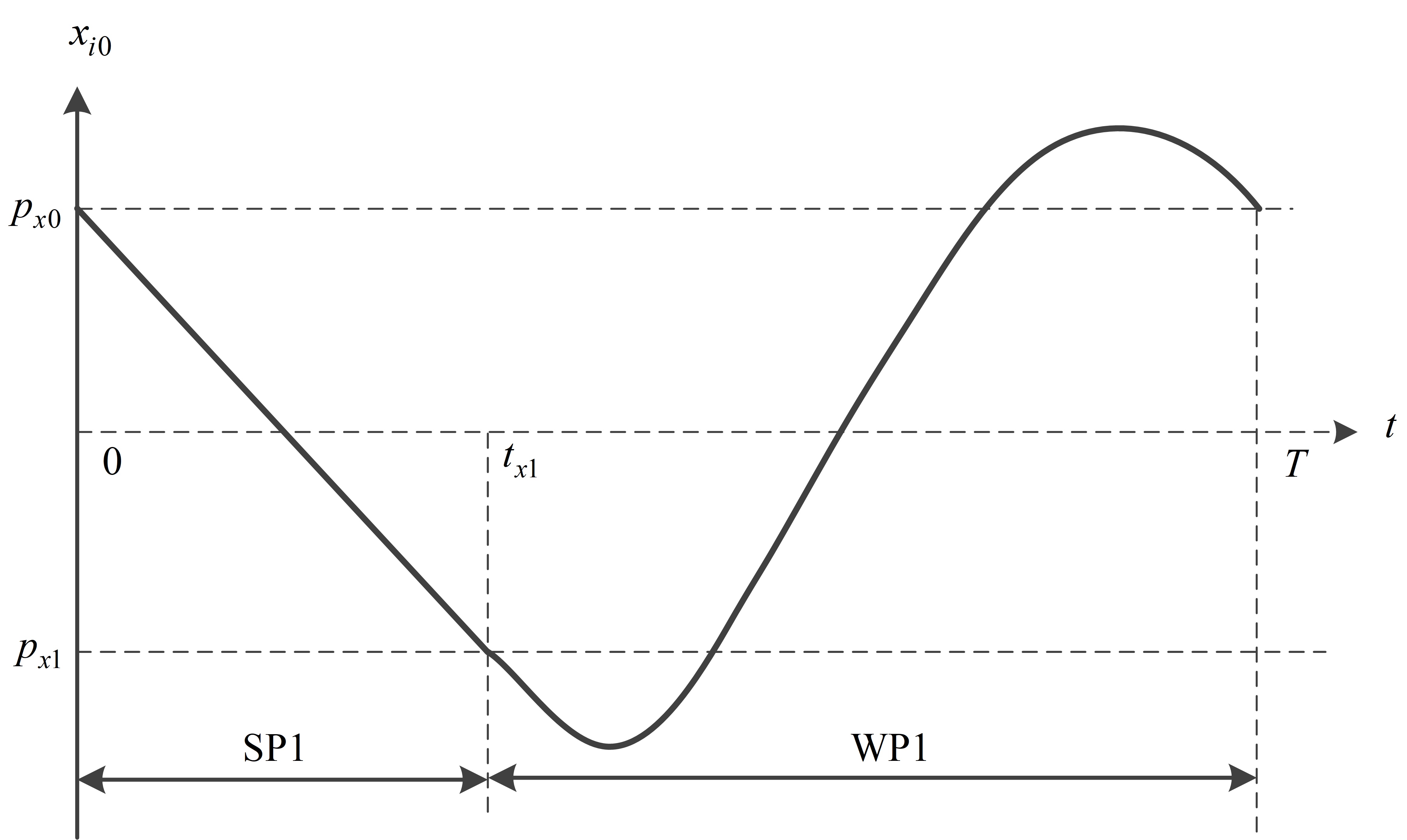}
\caption{Foot trajectory on the x-axis in a gait cycle}
\end{figure}

\subsubsection{Support foot planning}
The planning of the support foot is a main determinant of the moving speed of the robot. We generally expect the robot to move with a constant speed on the $x$-axis, so the trajectory of the support foot on the $x$-axis is planned as follows:

\begin{equation}
p_x(t)=p_{x0}-\int\nolimits_0^{t} v_x dt
\end{equation}
where $p_{x0}$ is the initial position of the support phase on the $x$-axis, which can be obtained by the joint angles' feedback combined with forward kinematics.

\subsubsection{Swing foot planning}


The trajectory planning of the swing foot is determined by the position and speed of two endpoints, namely, $p_{x1},p_{x2},v_{x1},v_{x2}$. To ensure the continuity of the speed and position of the foot trajectory, $p_{x1}$ and $v_{x1}$ are determined according to the position and speed of the end point in the support phase. To weaken the horizontal impact when the foot lands, $v_{x2}$ is determined by the desired forward speed of the robot.

\begin{equation}
\left\{
\begin{array}{l}
p_{x1}=p_{x0}+\int\nolimits_0^{t_{x1}} v_x dt\\
v_{x1}=-v_x\\
v_{x2}=-v_{x}\\
\end{array}
\right.
\end{equation}

\begin{figure}[htbp]
\centering
\includegraphics[width=3.0in]{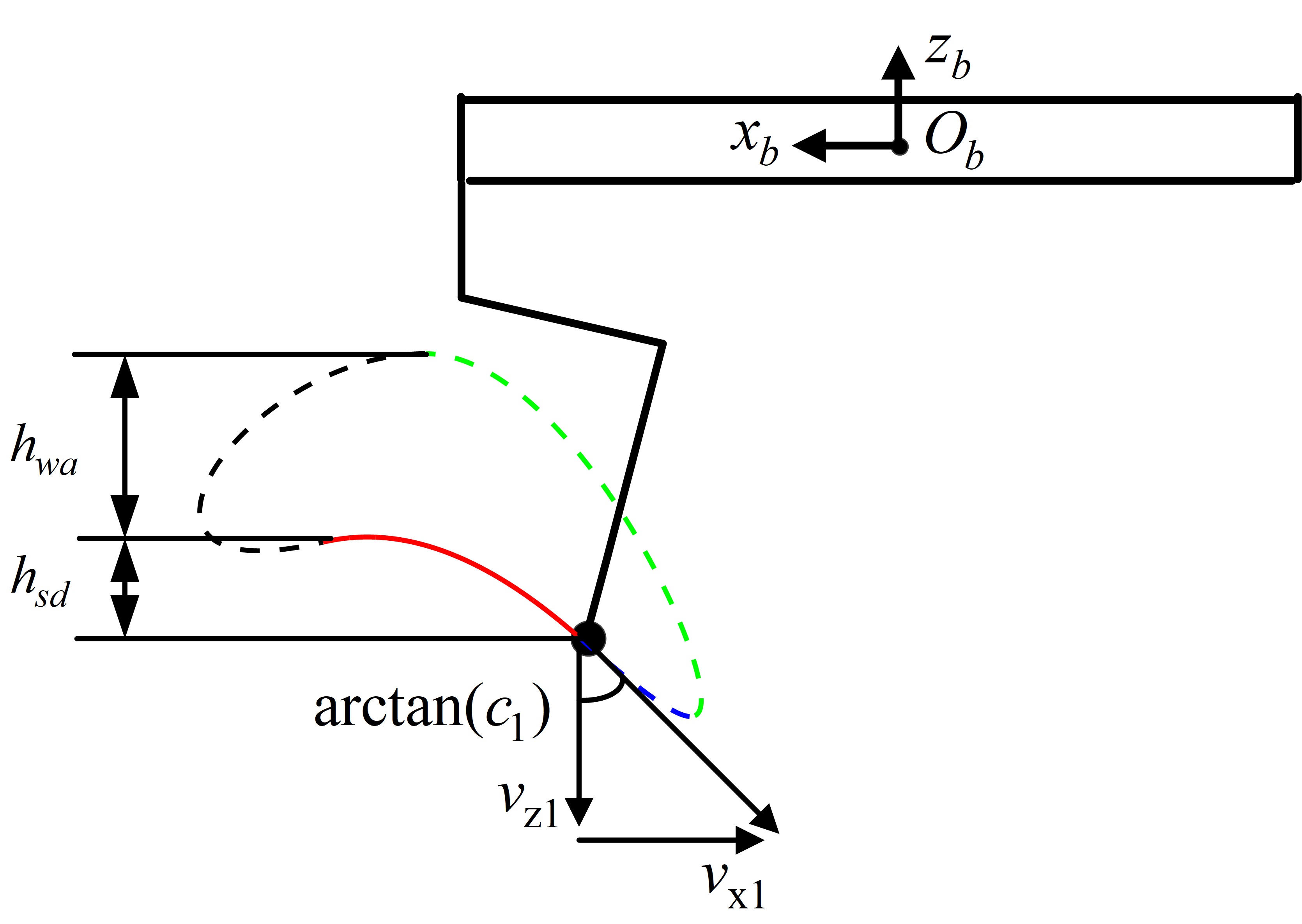}
\caption{Foot trajectory in the plane of $xoz$}
\end{figure}


\indent $p_{x2}$ is the landing position of the swing foot, which is selected to control the trunk speed. In the movement of the robot, due to various factors such as the foot sliding and the posture fluctuation of the trunk, the actual speed of the robot will deviate from what was planned. According to Raibert's research, since the landing position of the swinging foot has an effect on the speed of the trunk, the landing position selection can be used to adjust the actual speed of the robot and reduce its error from the desired speed.

\begin{equation}
p_{x2}=v_{bf,x}t_{x1}+k_{comp,x}(v_{bf,x}-v_{x})
\end{equation}
$v_{bf,x}$ is the trunk speed estimation on the $x$-axis, and $k_{comp,x}$ is the speed regulation constant on $x$-axis.

The foot trajectory planning on the $x$-axis is shown in Fig. 7. The trajectory planning of the left fore foot of the robot in the plane of ${xoz}$ is shown in Fig. 8.


\subsection{Directional Control}
The robot does not always need to move in a straight line, and sometimes, it needs to change its direction to avoid obstacles and move towards its destination. The rotation matrix can be used to rotate the foot trajectory of the robot on the horizontal plane to obtain the new foot trajectory of the robot that makes the robot rotate along the z-axis. Specify the desired heading angular velocity $\omega_z$; then, transform the robot's foot trajectory as follows:
\begin{equation}
{\bm{ \mathrm{p} }_p^S}^{'}=
\begin{pmatrix} 
	cos(\omega_zt)&-sin(\omega_zt)&0
	\\sin(\omega_zt)&cos(\omega_zt)&0
	\\0&0&1 
	\end{pmatrix}  
{\bm{\mathrm p}_p^S}
\end{equation}
\begin{equation}
{\bm{\mathrm p}_p^W}^{'}=
\begin{pmatrix} 
	cos(-\omega_zt)&-sin(-\omega_zt)&0
	\\sin(-\omega_zt)&cos(-\omega_zt)&0
	\\0&0&1 
	\end{pmatrix}  
{\bm{\mathrm p}_p^W}
\end{equation}
${\bm{\mathrm p}_p}=({p_x},{p_y},{p_z})$ is the foot position that is obtained from the foot trajectory planning in the previous two sections.\\


\section{Real-time Control}
\indent To enhance the robustness of robot trotting and its adaptability to a complex terrain, this paper uses some real-time control strategies to control the posture of the robot and weaken the impact of the interaction between the robot and the environment through the active compliance control of the robot. This chapter describes in detail the posture control method, the active compliant control method, and the selection of related parameters and, finally, provides the control structure diagram of the robot.

\subsection{Posture Control}
The quadruped robot may encounter many unexpected disturbances when moving, which may cause the robot body posture to fluctuate greatly or even directly cause the robot to fall down. To enhance the stability of robots in trotting, it is necessary to employ the dynamic control strategy to control the robot's posture.

\subsubsection{Virtual torque}

\begin{figure}[htbp]
\centering
\includegraphics[width=3.0in]{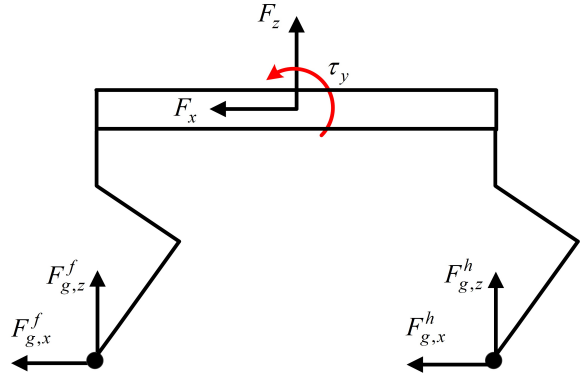}
\caption{Posture control by virtual torque}
\end{figure}


Since the mass of the legs of the robot is smaller than that of the robot trunk, the influence of the swinging legs on the posture of the robot can be ignored. The robot posture control can be realized by controlling the action of the support leg. Fig. 9 shows the lateral view of the robot when it is in the support phase of the flying trot gait. The robot is affected by the ground friction force along the $x$-axis and the ground support force along the $z$-axis. These forces are equivalent to two virtual forces and one virtual torque acting on the CoM of the trunk.

\begin{equation}
\tau_y=-h_{CoM} (F_{g,x}^{f}+F_{g,x}^{h} )-\dfrac{L}{2}(F_{g,z}^{f}-F_{g,z}^{h})
\end{equation}

\indent $h_ {CoM}$ is the height of the trunk CoM from the ground. $F_{g,x}^{f}$, $F_{g,z}^{f}$, $F_{g,x}^{h}$ and $F_{g,z}^{h}$ are the components of the ground forces acting on the front and hind feet in the $x$-axis and $z$-axis directions, respectively. $F_{x}$ and $F_{z}$ are the virtual forces acting on the trunk CoM in the $x$-axis and $z$-axis directions, and $\tau_y$ is the virtual torque acting on the trunk CoM in the $y$-axis direction.

According to the above analysis, the posture of the robot can be controlled by using virtual torque. The corresponding PD controller is as follows:
\begin{equation}
\tau_y=k_{p\tau,y} (\phi_{d,y}-\phi_{f,y} )+k_{d\tau,y} (\dot{\phi}_{d,y}-\dot{\phi}_{f,y} )
\end{equation}

Since $h_{CoM}$ changes very little, $\tau_y$ is proportional to $F_{g,x}^{f}$ and $F_{g,x}^{h}$. Assuming that no sliding occurs between the ground and the support foot, $F_{g,x}^{f}$ and $F_{g,x}^{h}$ are proportional to the acceleration of the foot relative to the trunk, $\ddot{p}_{adj,x}^{f}$ and $\ddot{p}_{adj,x}^{h}$, respectively. Therefore, we can control the trunk posture with $\ddot{p}_{adj,x}^{f}$ and $\ddot{p}_{adj,x}^{h}$. Specify $\ddot{p}_{adj,x}^{f}$=$\ddot{p}_{adj,x}^{h}$; then, the posture controller used in this paper is as follows:

\begin{equation}
\ddot{p}_{adj,x}^{f}=\ddot{p}_{adj,x}^{h}=k_{p\tau,y} (\phi_{d,y}-\phi_{f,y} )+k_{d\tau,y} (\dot{\phi}_{d,y}-\dot{\phi}_{f,y} )
\end{equation}

The desired foot position can be obtained by the foot planning position and the adjusting acceleration:
\begin{equation}
\bm{\mathrm{p}}_{d}={\bm{\mathrm{p}}_p}^{'}+\iint\ddot{\bm{\mathrm{p}}}_{adj}dt
\end{equation}



\subsubsection{Touch on obstacles}
The swing foot may touch on obstacles early in the gait cycle, such that the posture of robot will fluctuate rapidly and the robot may even fall down if the trajectory of the swing foot is still tracked in this situation. To reduce the fluctuations of the posture, the action of the swing foot on the $z$-axis should be stopped.   

\subsection{CoM regulation}
When the projection of the robot's center of mass on the ground is at the center of the support line, the motion of the robot will be more stable and natural. We will discuss below.
\subsubsection{CoM observation}
To simplify the description, we assume that the robot is trotting in place. Suppose the center of mass of the robot is behind the support line. The ZMP(zero momentum point) and CoM have the same coordinates on x-axis since robot will not move on x-axis, as shown in Fig. 10.  When the left front leg swings, the robot body will move to the left to maintain balance \cite{Biped_walking_pattern_generation}. When the right front leg swings, the robot body will move to the right. This will lead to the difference of the front and rear foot touchdown position in y axis, and even crossed legs.
\begin{figure}[htbp]
\centering
\includegraphics[width=4.0in]{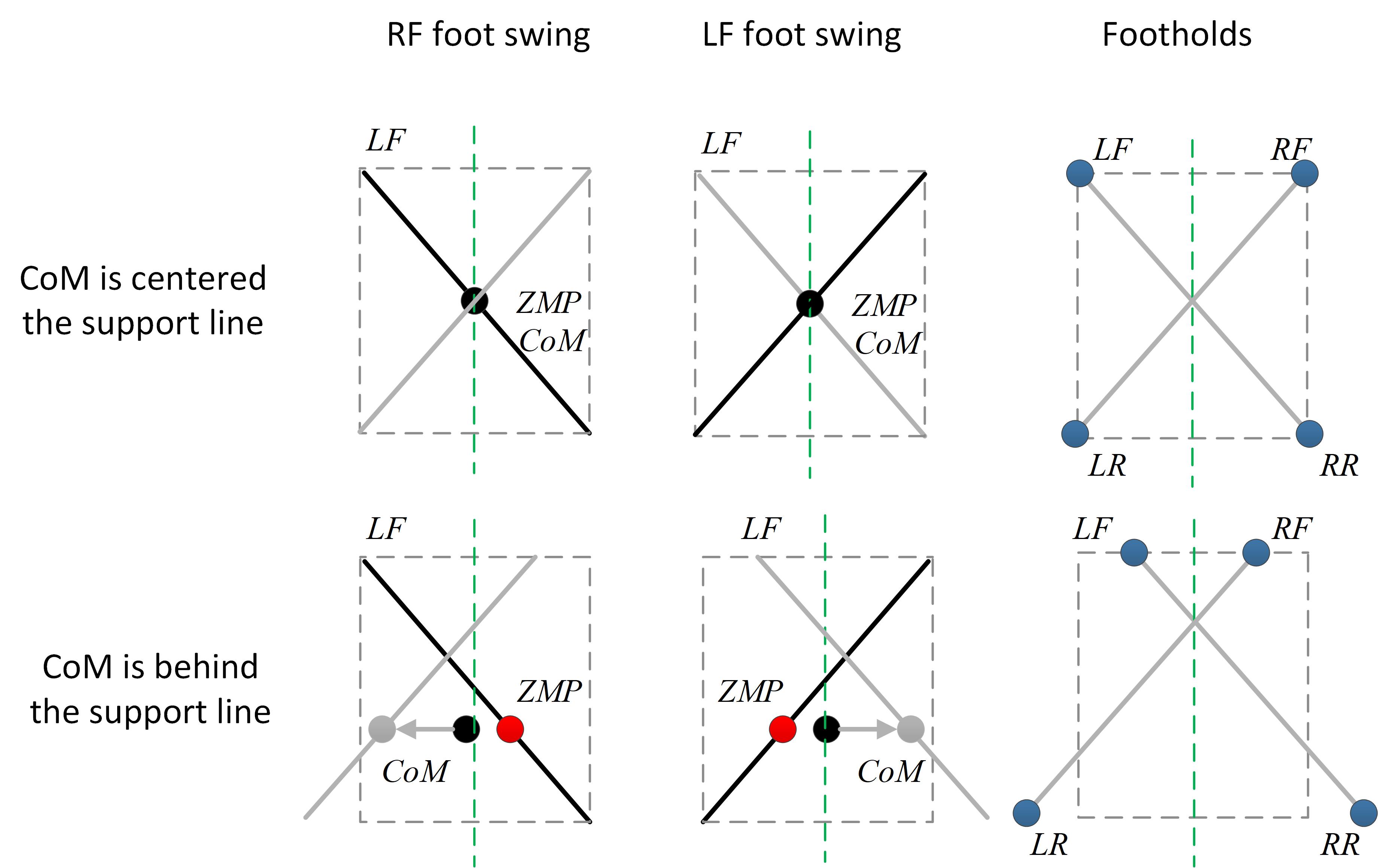}
\caption{CoM is behind the support line. Grey dashed line is body of robot, green dashed line is 
symmetrical line of body. Black solid line is current support line, grey solid line is next support line, red point is ZMP, black point is CoM, blue point is the latest touchdown position for all foot. LF, RF, LR, RR is abbreviations of left front, right front, left rear and right rear.}

\end{figure}

\subsubsection{Adjust touchdown position}
Therefore, we can judge whether the center of mass is in front of or behind the support line by the symmetry of the touchdown position of the front and rear foots in y axis. Further, we can control the center of the CoM on the support line by modifying the x-aixs foot touchdown position from equation(12):

\begin{equation}
p_{x2}=v_{bf,x}t_{x1}+k_{comp,x}(v_{bf,x}-v_{x})+k_{CoM}(p_{lr,y}-p_{lf,y}+p_{rf,y}-p_{rr,y})
\end{equation}
where $p_{lf,y},p_{rf,y},p_{lr,y},p_{rr,y}$ is current position in y-axis for left front foot, right front foot, left rear foot and right rear foot.

\subsection{Leg Compliance}
When a quadruped robot is running, the position of its trunk on the z-axis will fluctuate greatly, which will produce a great impact when its foot touches the ground. In the third section of this paper, the impact is partially buffered by planning the buffering action in the foot trajectory. To further enhance the buffering effect, the active compliance based on floating base inverse dynamics and the leg virtual model control can be used. However, due to the need of an accurate identification of the leg inertia and an accurate estimation of the foot force in floating base inverse dynamics, which are difficult tasks, this paper employs the leg virtual model control for active compliance control, as shown in Fig. 11.


\begin{equation}
\left\{
\begin{array}{l}
F_{vir,x}^{i}=k_{p,x}^{i} (p_{d,x}^{i}-p_{f,x}^{i} )+k_{d,x}^{i} (\dot{p}_{d,x}^{i}-\dot{p}_{f,x}^{i} )\\
F_{vir,y}^{i}=k_{p,y}^{i} (p_{d,y}^{i}-p_{f,y}^{i} )+k_{d,y}^{i} (\dot{p}_{d,y}^{i}-\dot{p}_{f,y}^{i} )\\
F_{vir,z}^{i}=k_{p,z}^{i} (p_{d,z}^{i}-p_{f,z}^{i} )+k_{d,z}^{i} (\dot{p}_{d,z}^{i}-\dot{p}_{f,z}^{i} )\\
\end{array}
\right.
\end{equation}


In the formula, $x_{d,i}$, $y_{d,i}$ and $z_{d,i}$ are the desired position of foot $i$, while $x_{a,i}$, $y_{a,i}$ and $z_{a,i}$ are the actual position of foot $i$. $f_{x, i}$, $f_{y,i}$, and $f_{z, i}$ are the virtual forces that will be applied to foot $i$.

\begin{figure}[htbp]
\centering
\includegraphics[width=3.5in]{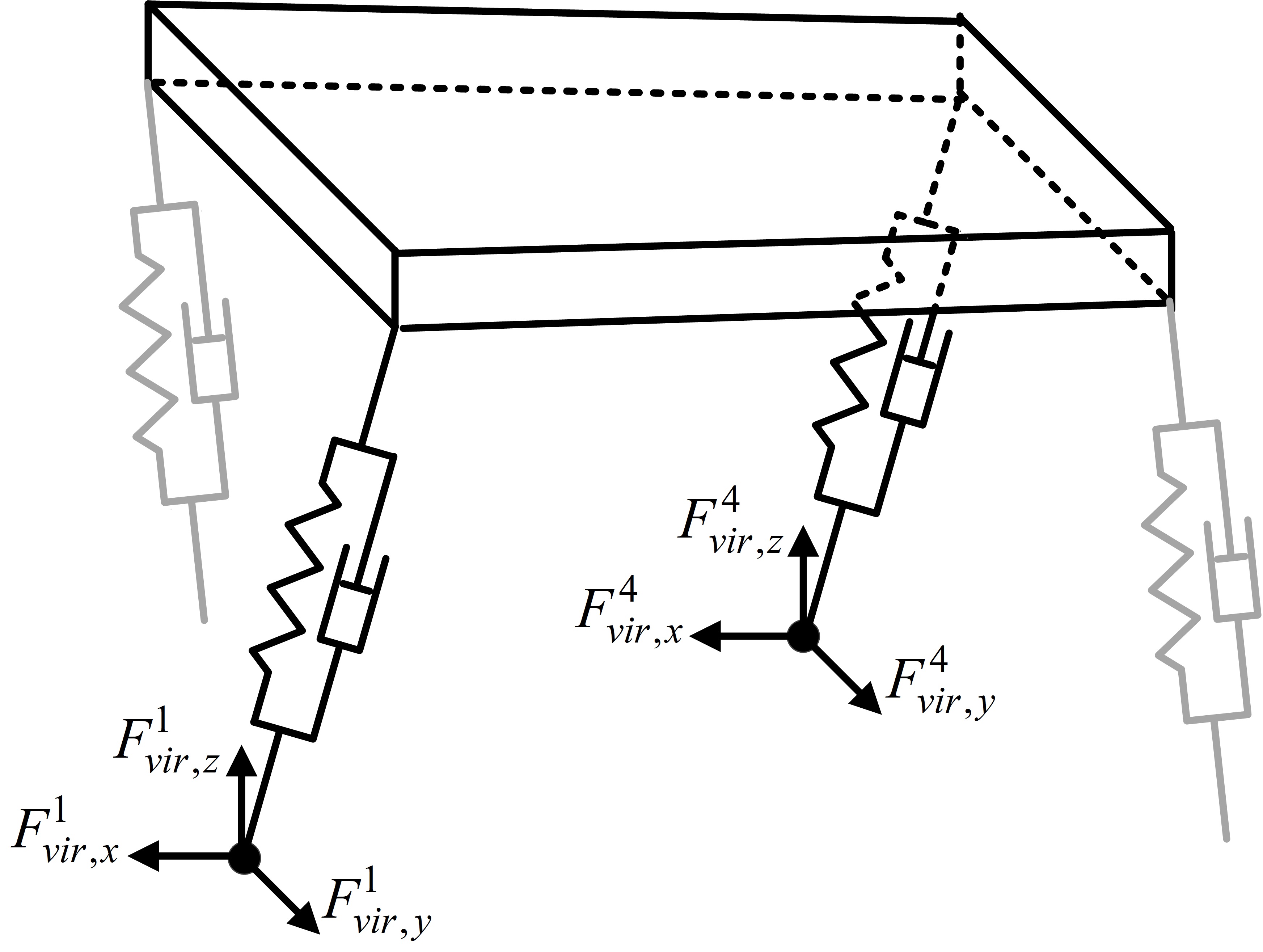}
\caption{The virtual impedance and damping of the leg}
\end{figure}

\subsubsection{Gravity compensation}

In the support phase, the two support feet bear the gravity of the whole robot. To ensure the accuracy of the position control of the robot foot on the $z$-axis, a relatively large $k_{p,z}^{i}$ is needed. However, when the $k_{p,z}^{i}$ is too large, it is easy to cause instability of the system. In this paper, the gravity compensation method is adopted to solve this problem. In the support phase, there are two feet on the ground, as shown in Fig. 9. The support force provided by foot $i$ is $F_{g,z}^{i}$, and the gravity compensation for foot $i$ is $F_{Comp}^{i}$; then, there is


\begin{equation}
\left\{
\begin{array}{l}
F_{g,z}^{i}=-(F_{vir,z}^{i} +F_{Comp}^{i}) \\
G=-\sum\limits_{i\in S}F_{g,z}^{i}\\
\end{array}
\right.
\end{equation}

\indent $G$ is the gravity of the robot, and $S$ represents the group of the support feet.



\subsubsection{The selection of \texorpdfstring{$k_{p,zi}$}{pdfstring} and \texorpdfstring{$k_{d,zi}$}{pdfstring}}
In the flying trot gait, the motion of the robot on the z-axis is in a state of periodic up and down oscillation, and the whole process includes the landing buffering, ejection from the ground and flying of the robot. The control of the robot on the z-axis, which is mainly affected by the control of the support leg, is particularly important because it directly determines whether the robot can achieve a stable flying trot gait. Since the selection of $k_{p,zi}$ and $k_{d,zi}$ of the support leg has a great influence on the control of the support leg, their selection should be decided upon with caution.


The leg virtual model of the robot can be equivalent to a spring-damped system with an oscillation period determined by $k_{p,zi}$ and $k_{d,zi}$. If $k_{p,zi}$ and $k_{d,zi}$ are not suitable, the step period of the robot will be disturbed and even result in the oscillation divergence of the robot movement on the $z$-axis, eventually causing the instability of the robot. To ensure that the oscillation of the spring-damping system is in sync with the planning oscillation of the flying trot on the $z$-axis, when selecting $k_{p,zi}$ and $k_{d,zi}$, the oscillation period of the spring-damping system is guaranteed to be equal to twice that of $\Delta t_{z1}$, so the spring-damping system can complete two stages of compression and ejection in $\Delta t_{z1}$.


\begin{equation}
\left\{
\begin{array}{l}
\omega_n = \sqrt{\frac{k_{p,z}^{i}}{m_s}}\\
\zeta=\frac{k_{d,z}^{i}}{2\sqrt{k_{p,z}^{i}m_s}}\\
\omega_d=\omega_n\sqrt{1-\zeta^2}\\
\Delta t_{z1}=\frac{\pi}{\omega_d}\\
\end{array}
\right.
\end{equation}

\indent $m_s$ is half the mass of the robot, $\omega_n$ is the natural frequency of the spring oscillator in the undamped state, $\zeta$ is the damping ratio, and $\omega_d$ is the oscillation frequency of the spring-damped system.

\subsubsection{Map foot force to joint torque}
Since the foot force cannot be directly controlled in actual control, it is necessary to map the foot force to the joint torque using the Jacobi matrix.



\begin{equation}
\bm{\mathrm \tau}_d^i =\bm{\mathrm{J}}^{T}  \bm{\mathrm F}_d^i 
\end{equation}

\indent $\bm{\tau_i}$ is the desired joint torque vector, and $\bm{\mathrm F}_d^i=(F_{vir,x}^{i} ,F_{vir,y}^{i},F_{vir,z}^{i} +F_{Comp}^{i})$ is the desired foot force vector. $\bm{\mathrm J}$ is the Jacobi matrix.



The control structure of the flying trot gait control method proposed in this paper is shown in Fig. 12.

\begin{figure}[htbp]
\includegraphics[width=3.7in,center]{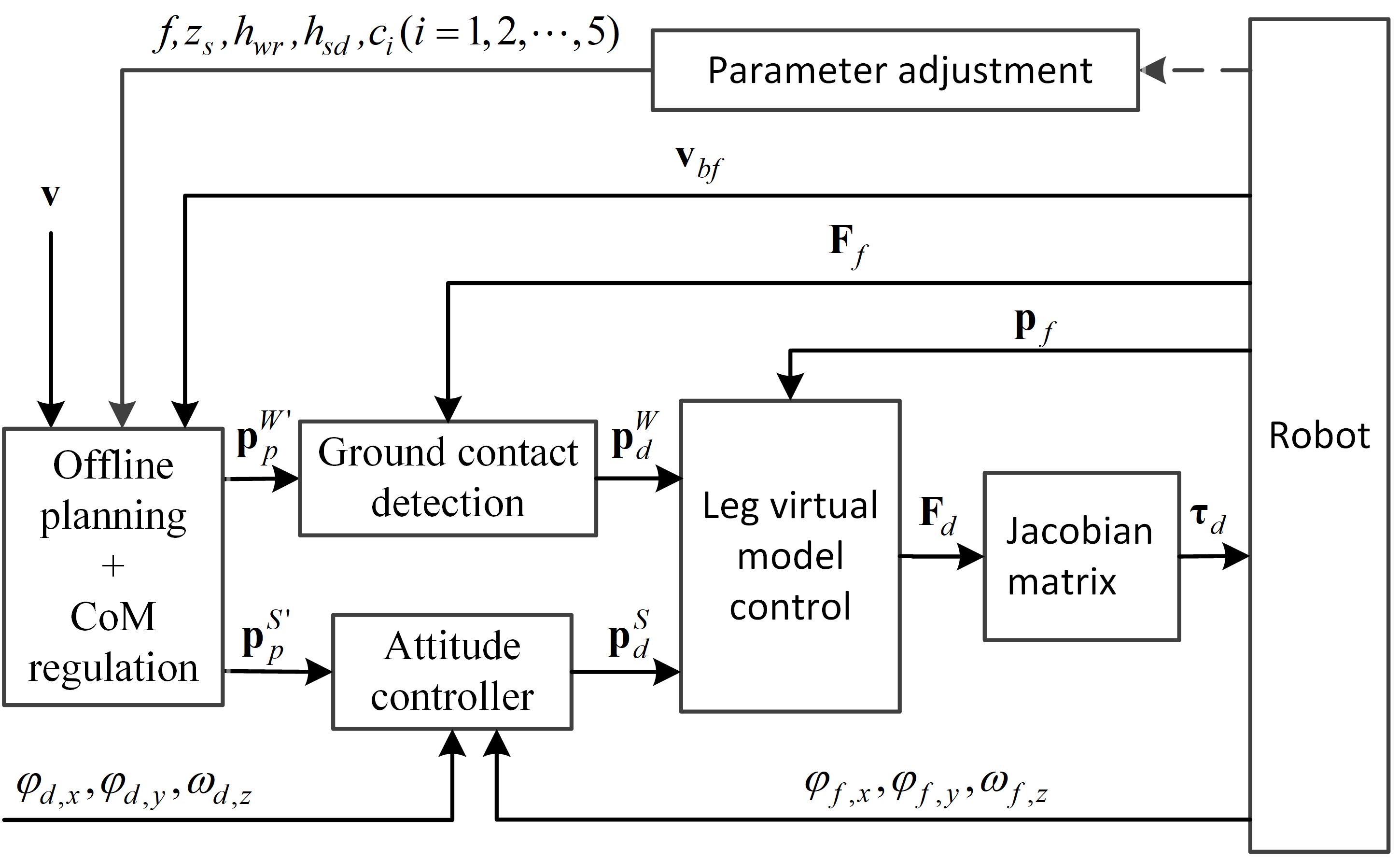}
\caption{Control structure of the flying trot gait}
\end{figure}

\section{Simulation and Experiment}

\indent To verify the feasibility of the method proposed in this paper, the control method proposed in this paper was verified in the Webots simulation environment employing the Billy robot. The effect of the flying control, speed control, posture control and balance control are tested in the simulation and experiment.

\subsection{Simulation}
%

The simulation was carried out in the Webots simulation software. The screenshot of the simulation process was taken, and the sensor data generated in the simulation process were drawn with MATLAB to observe the control effect of the robot. In the simulation, quadruped robots can quickly stabilize from an external disturbance and can run at high speeds. The simulation results prove that the control method proposed in this paper can drive the robot to run with a flying trot gait.

\subsubsection{Balance control}
The flying trot gait of a quadruped robot is not static and stable because the robot has no more than two support legs forming a support line in the flying trot. In the process of running, the robot's trunk may be disturbed by external disturbance. Studies have shown that the external disturbance on the side of a quadruped robot is more likely to make the robot unstable or even fall down compared to the external disturbance on the front. \\
\indent During the running process of the robot, a 1.5-kg pendulum is used to impact the side of the robot to simulate the possible disturbance acting on the robot in the real environment. The moment before the pendulum impacts the robot, its momentum is approximately 2.1 kg m/s. The simulation snapshots in Webots are shown in Fig. 13, and the sensor data generated in the simulation are shown in Fig. 14, in which the red line stands for the desired value, and the green line stands for the actual value. After the impact, the robot trunk's roll angle reached the maximum after 1 second, which was 0.05 rad. After the adjustment of the posture controller, the robot body stabilized after 0.8 seconds, and the roll angle stabilized within $\pm 0.02$ rad.


It can be seen that the pitch angle also fluctuates after the impact. This is because the robot will rotate around the support line formed by the support feet after impact, which will cause both the roll angle and the pitch angle to fluctuate.

\begin{figure*}[htbp]
\centering
\includegraphics[width=5.0in]{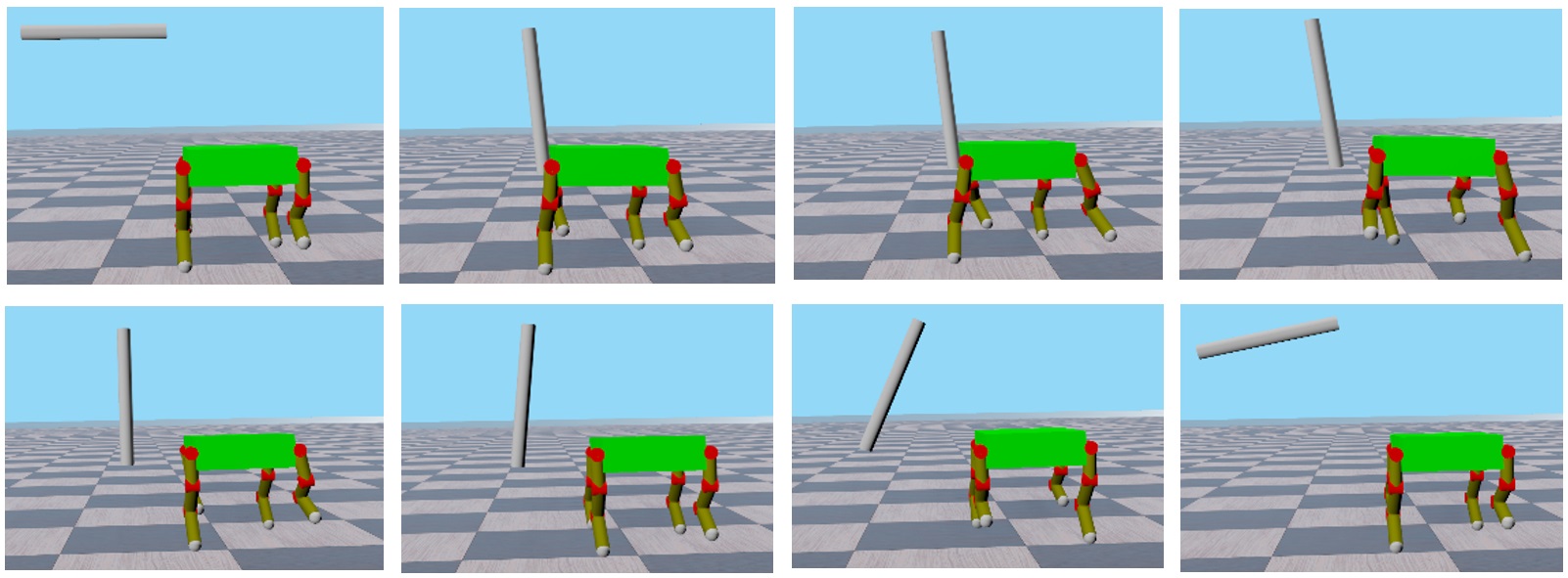}
\caption{Balance control simulation in Webots}
\end{figure*}

\begin{figure}[htbp]
\centering
\includegraphics[width=3.5in]{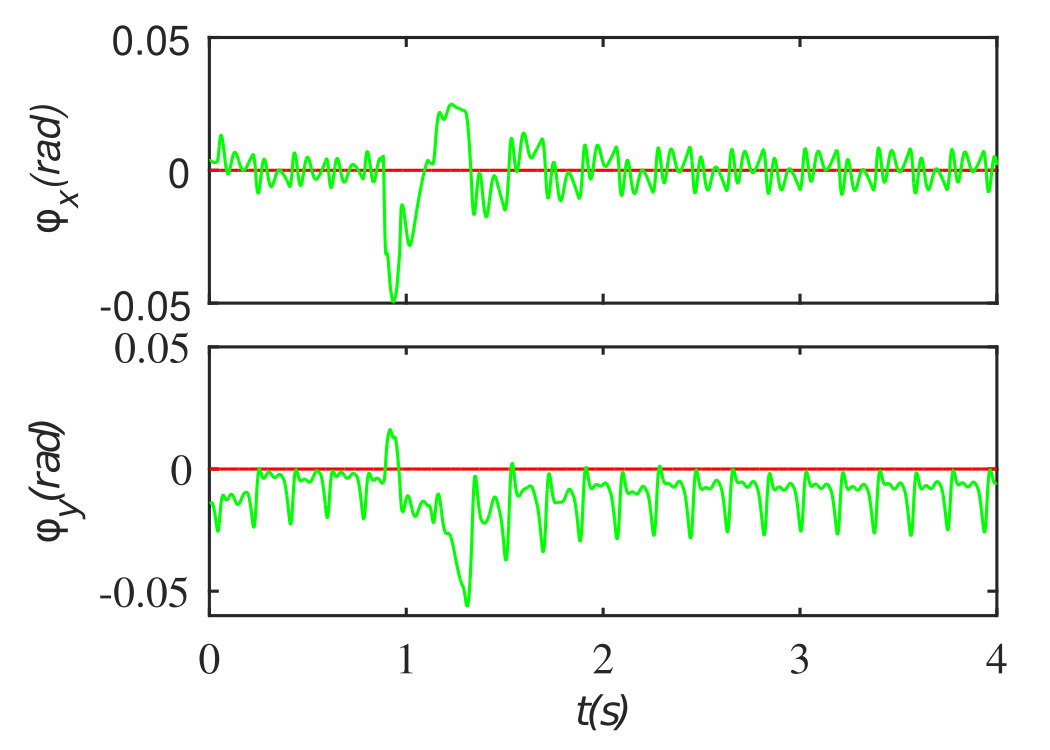}
\caption{The posture data in the balance control simulation}
\end{figure}



\subsubsection{Running control}
One of the great advantages of a flying trot over a walking trot is that it can move the robot at a high speed. To verify the overall effect of running control with the flying trot gait proposed in this paper, a running control simulation was carried out.

The control parameters are set as $f = 2.9$ Hz, $v_x =0.0 \sim 1.0$ m/s, $z_s = -0.14 $ m, $h_{wa} = 0.04$ m, $h_{sd} = 0.005$ m, $c_1 = 4.0$, $c_2 = -0.1$, $c_3 = -3.0$, $c_4 = -4.5$, and $c_5 = 0.2$. $v_x$ increased from 0.0 m/s to 1.0 m/s and then decreased from 1.0 m/s to 0.0 m/s during the simulation to test the effect of speed tracking. The snapshots of the quadruped robot running are shown in Fig. 15, and the sensor data generated in the process of simulation are shown in Fig. 16, in which the red line stands for the desired value, and the green line stands for the actual value. It can be seen that there is a quadruped flying phase in the running process. The desired forward speed can be tracked by the robot rapidly with a deviation within $\pm 0.2$ m/s. The posture fluctuation is within $\pm 0.06$ rad. 



The CoM height of the robot presents the oscillation characteristics of up and down, corresponding to the four stages of the landing cushion, ejection from the ground, ascending and descending, which coincides with the movement of the SLIP model. The reason why the forward speed fluctuates is that, due to the air resistance, the speed of the robot will drop when it is in the flying phase, and when it is in the support phase, the support foot will promote the trunk to accelerate forward to make the actual speed of the trunk closer to the desired speed. There is a difference between the average value of the actual speed of the trunk and the desired speed, mainly because the posture control algorithm will adjust the foot trajectory planning of the support foot, which causes a disturbance to the speed control of the trunk.

\begin{figure*}[htbp]
\centering
\includegraphics[width=5.0in]{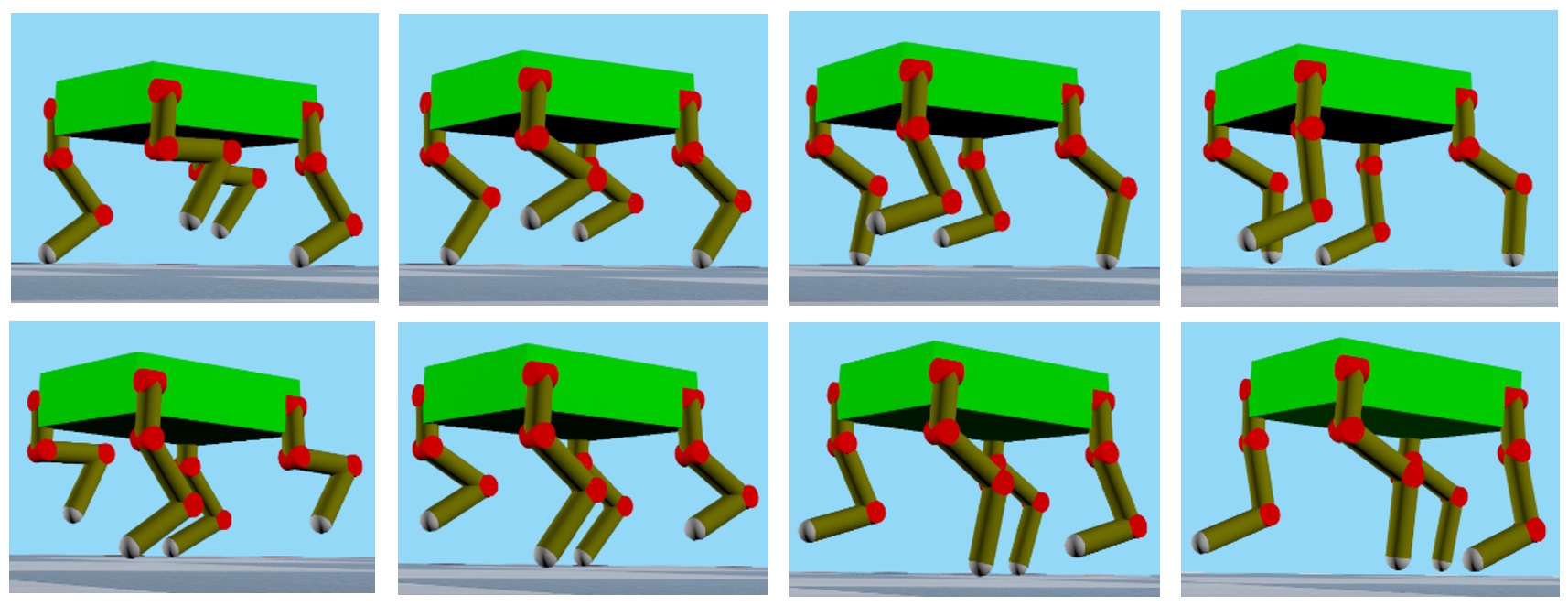}
\caption{The snapshots of the running control simulation}
\end{figure*}

\begin{figure}[htbp]
\centering
\includegraphics[width=3.0in]{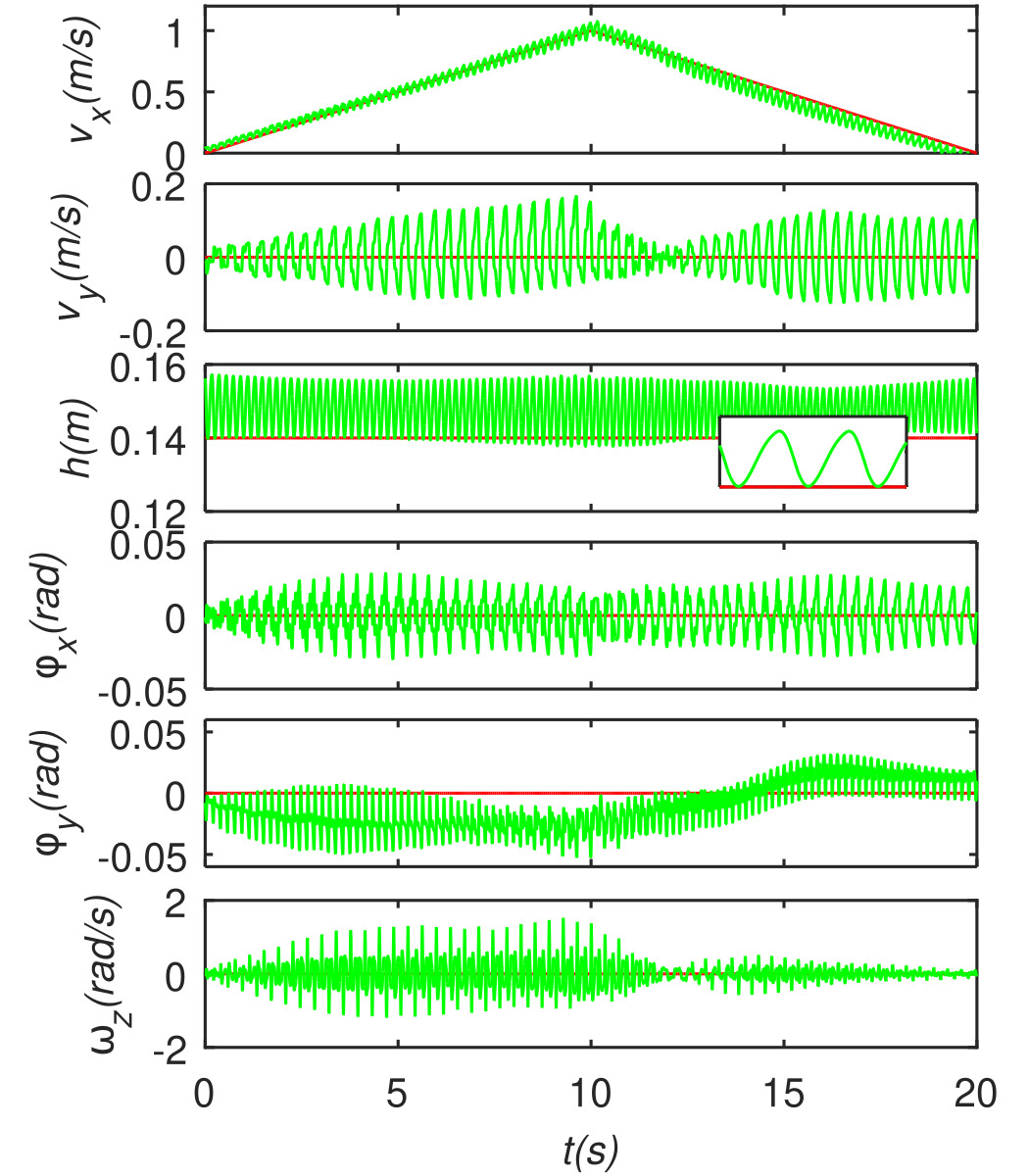}
\caption{The results of the running control simulation}
\end{figure}

\indent The CoM trajectory of the robot trunk and the foot trajectory during the running control simulation of the quadruped robot is shown in Fig. 17.  

\begin{figure}[htbp]
\centering
\includegraphics[width=3.5in]{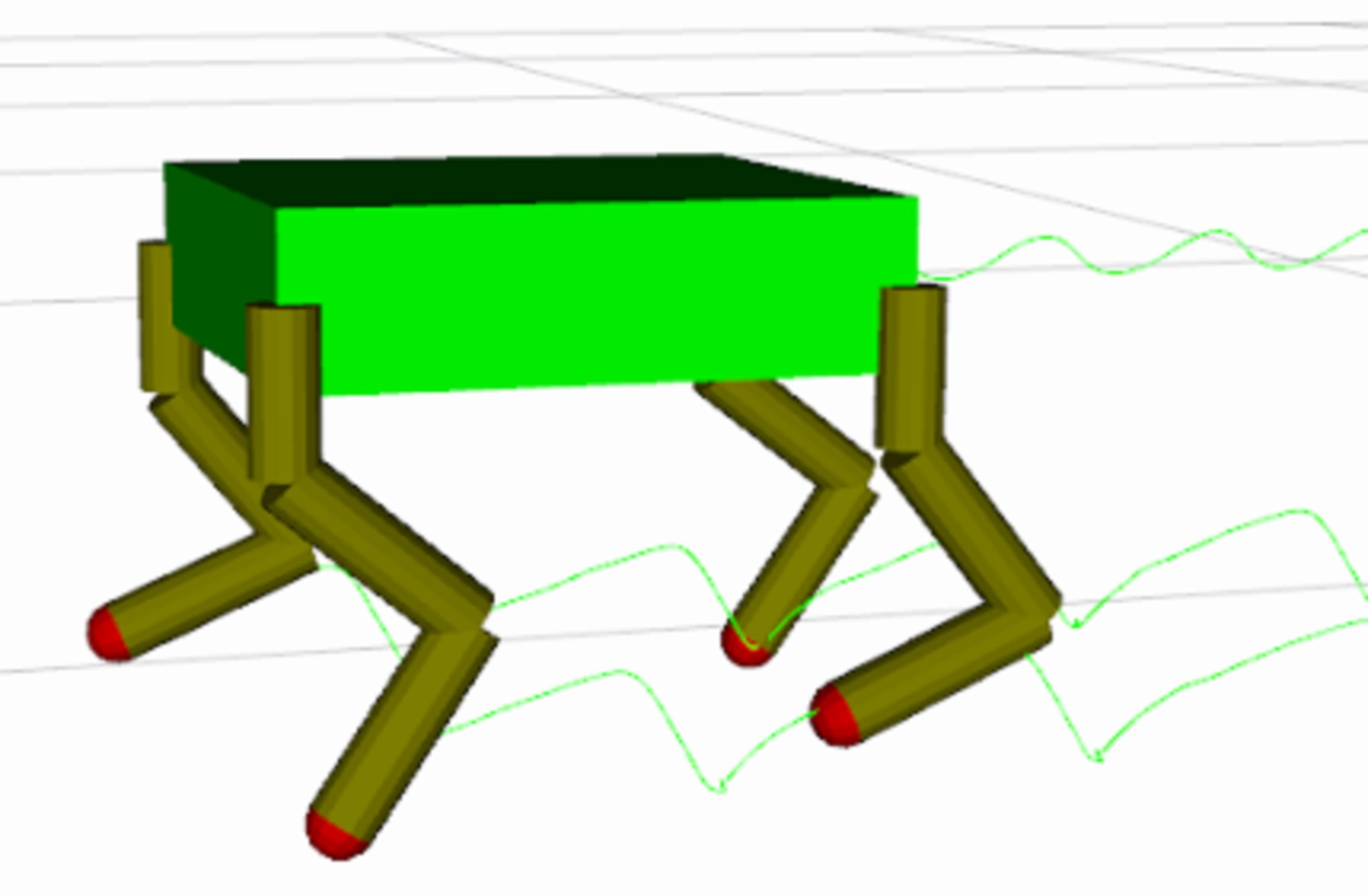}
\caption{Foot trajectory and body CoM trajectory in the flying trot}
\end{figure}

\subsection{Experiment}
In this paper, two experiments were carried out to verify the flying control and the speed control using Billy. Because there is no foot force sensor or joint torque sensor on Billy, the accurate time of touching the ground cannot be detected, so only the planning method proposed in this paper is used in the experiment, and no posture control, leg virtual model and other controls are used. The experiment was carried out on flat ground, and the process was recorded by a camera.


\subsubsection{Flying control}

The flying phase in the gait cycle is the obvious difference between the walking trot and the flying trot and is one of the factors that enable the flying trot to achieve rapid movement. To verify whether the control method proposed in this paper can control the robot to achieve flying in quadruped trotting, flying control experiments were carried out.

To clearly observe the situation of the foot leaving the ground when the robot is in the flying phase, the Billy robot with the black foot was selected in this experiment, and the ground in the experiment was milky white. The video snapshots of experiment are shown in Fig. 18. It can be seen that the diagonal legs of Billy eject from the ground alternately, and then, the robot is completely off the ground in the experiment. The ratio of the flying time of Billy in a gait cycle to the total gait cycle time is calculated, and the corresponding duty factor of the support phase is 0.31, as shown in Fig. 19.

%
%

\begin{figure*}[htbp]
\centering
\includegraphics[width=5.0in]{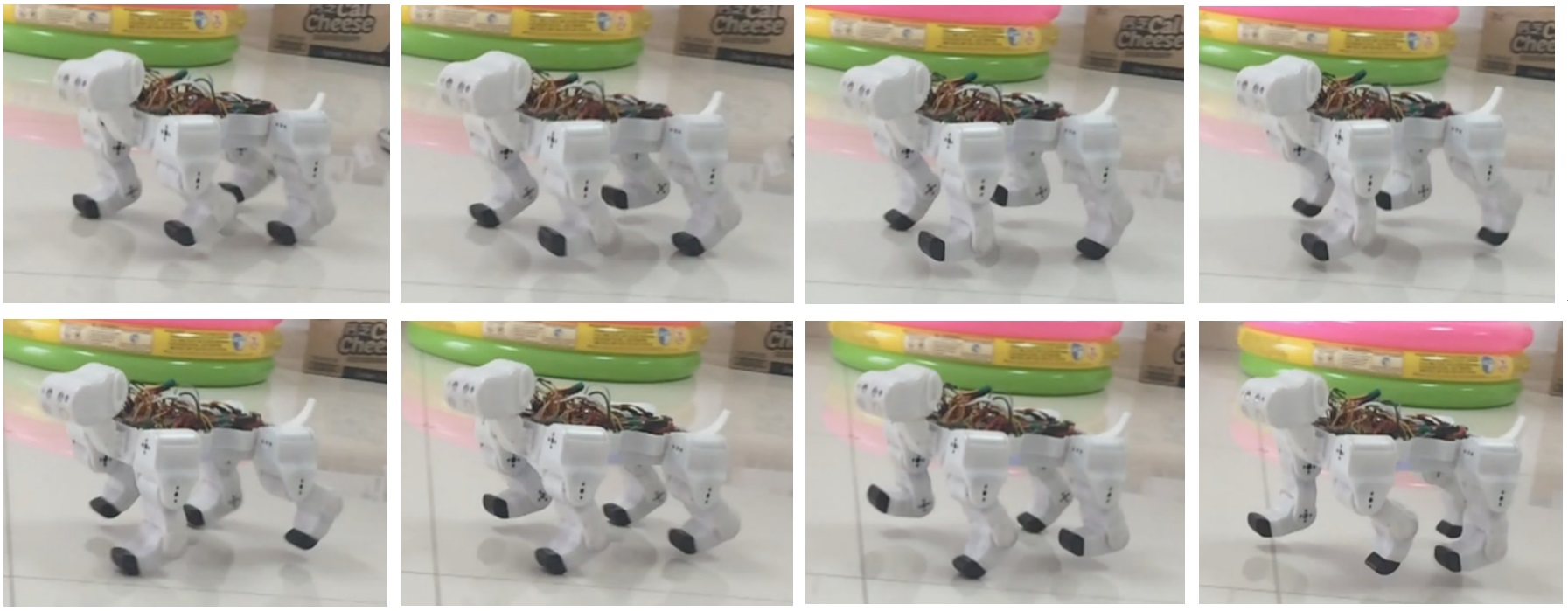}
\caption{Video snapshots in the flying control experiment}
\end{figure*}

\begin{figure}[htbp]
\centering
\includegraphics[width=3.5in]{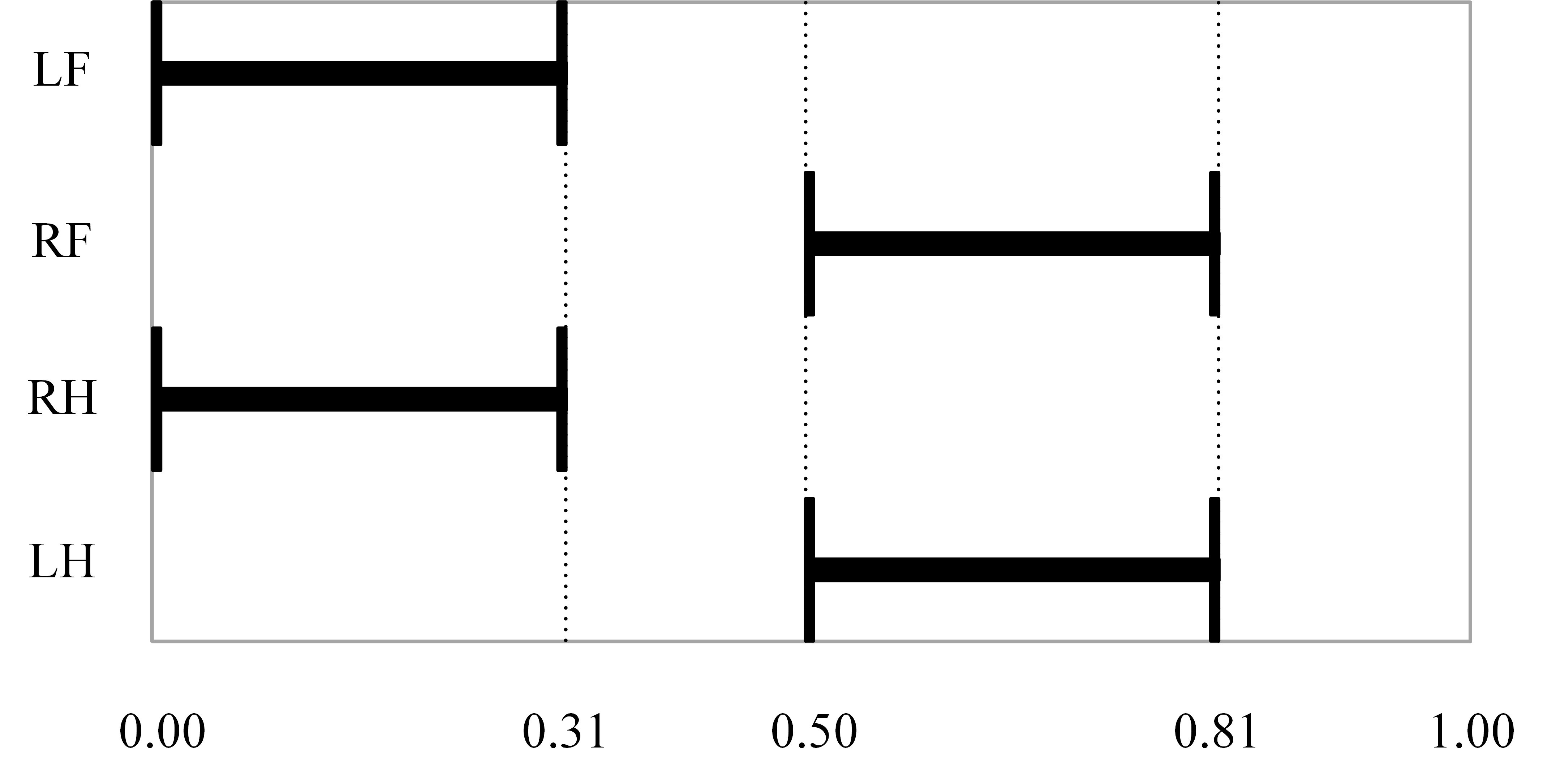}
\caption{Gait graph of Billy in the flying control experiment}
\end{figure}

%

\subsubsection{Speed control}
To verify the effect of speed control, the speed control experiment is carried out. Given $f = 2.8$ Hz, $v_x =0.8$ m/s, $z_s=-0.154$ m,$h_{wa} = 0.02$ m, $h_{sd} = 0.01$ m, $c_1=2.0$, $c_2=-0.1$, $c_3=-2.0$, $c_4=-0.1$, and $c_5=0.6$ in the experiment. There is a tape on the ground used to record the position of Billy in the experiment. The video snapshots of the experiment are shown in Fig. 20. As shown in the figure, the distance between each two small black points on the snapshot is 0.1 m, and the time interval between the two adjacent snapshots is 0.25 s. Therefore, the average speed of the quadruped robot while running can be calculated, which is approximately 0.81 m/s.

In the snapshots of the video, it can be seen that the quadruped robot can run stably with the flying trot gait, and there is no great posture fluctuation in the running process. In addition to the points mentioned in the simulation experiment, there is another reason for the difference between the actual average speed and the desired speed of the robot, which is the model error introduced in modelling Billy robot's aspherical foot with the spherical foot.

\begin{figure}[htbp]
\centering
\includegraphics[width=3.2in]{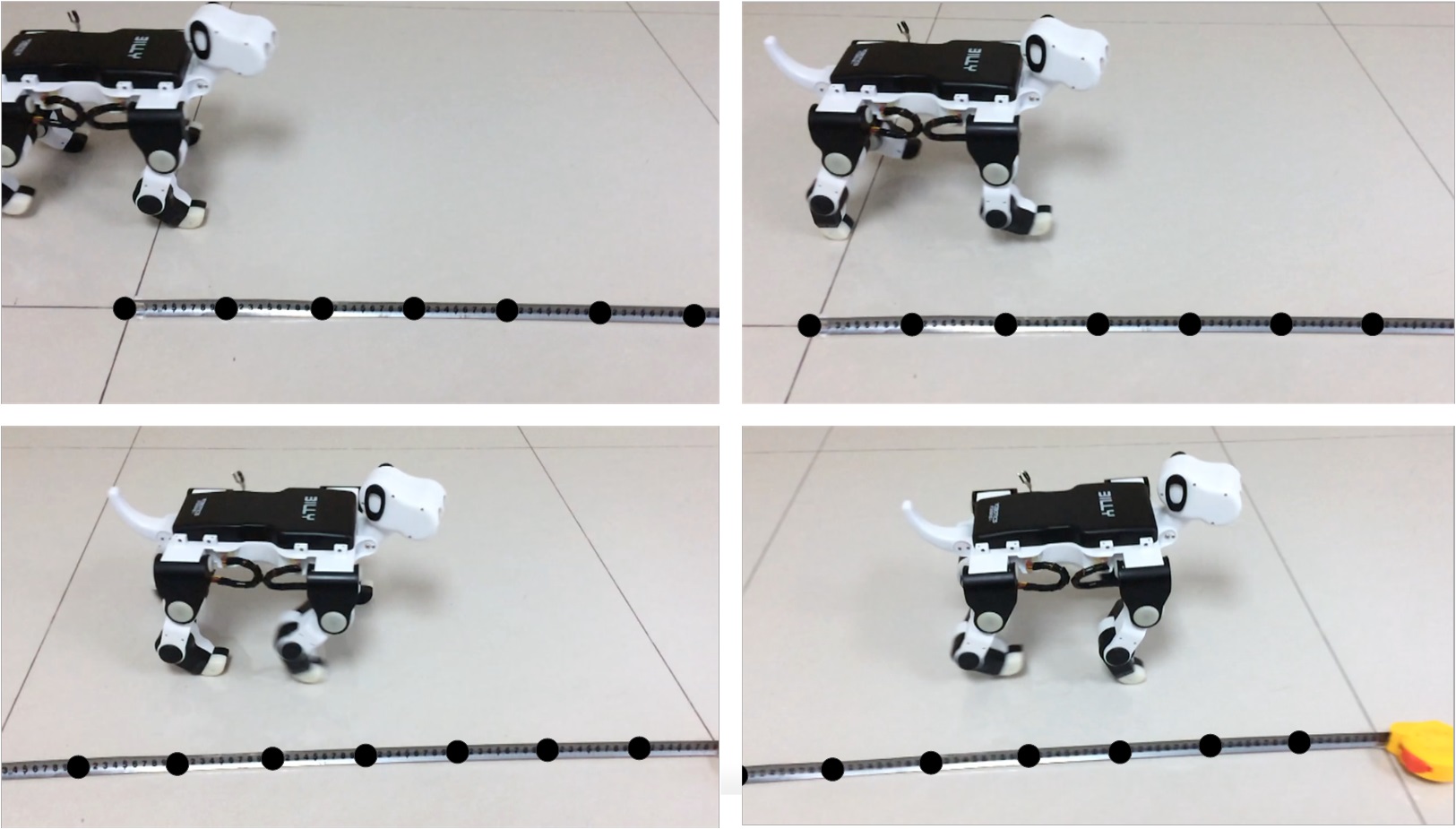}
\caption{Video snapshots in the speed control experiment}
\end{figure}

\section{Conclusion}
\indent In this paper, the control method based on the foot position planning was used to control the quadruped robot in 3 linear directions and 3 rotational directions to achieve quadruped running with the flying trot gait. The SLIP model's motion in the vertical direction is simulated by the planning method to realize the flying control of the robot. The forward and lateral speeds are controlled by the horizontal planning of the support foot and the selection of the landing position of the swinging foot. By adjusting the acceleration of the support foot in the horizontal direction, the posture of the robot trunk can be controlled. The method of adjusting the heading angular velocity with the support foot is also given.


The main advantages of the method presented in this paper are as follows. In the planning of the flying trot, several control parameters are proposed, which can be flexibly adjusted according to the actual situation to achieve the flying trot gait in various situations. According to the planning, the touchdown time of the robot can be predicted to make a buffering action in advance, which can reduce the impact from the touchdown. The posture of the trunk can be controlled by adjusting the acceleration of the foot. Compared to the method of controlling the posture of the trunk by adjusting the speed of the foot, the sudden change in the acceleration of the foot will not occur, which is conducive to the stable running. By means of the active compliance control of the leg virtual model, the impact of touchdown can be further reduced and the anti-disturbance ability of the system can be improved. By using the idea of resonance, the selection method of the virtual impedance and virtual damping in the leg virtual model is given, which makes the vibration period of the leg virtual spring-damping system consistent with the support phase time of the flying trot and coordinates the movement of the virtual spring-damping system and gait planning.

In the future, we will verify the proposed method on a larger quadruped robot platform. The method of simplifying the process of adjusting the robot control parameters will be studied based on online learning.

\section{Acknowledgements}

This work was supported by the National Key R \& D Program of China [grant numbers 2017YFC0806505, 2017YFB1302400]; the National Natural Science Foundation of China [grant number 9174821].

\bibliography{elsarticle-template}

\begin{thebibliography}{10}
\expandafter\ifx\csname url\endcsname\relax
  \def\url#1{\texttt{#1}}\fi
\expandafter\ifx\csname urlprefix\endcsname\relax\def\urlprefix{URL }\fi
\expandafter\ifx\csname href\endcsname\relax
  \def\href#1#2{#2} \def\path#1{#1}\fi

\bibitem{Legged_robots_that_balance}
M.~Raibert, Legged robots that balance, MIT press, 1986.

\bibitem{Quadruped_Hopping_in_legged_systems}
M.~Raibert, Hopping in legged systems—modeling and simulation for the
  two-dimensional one-legged case, IEEE Transactions on Systems, Man, and
  Cybernetics~(3) (1984) 451--463.

\bibitem{Kolt_System_design}
J.~Nichol, S.~Singh, K.~Waldron, L.~P. Iii, D.~Orin, System design of a
  quadrupedal galloping machine, The International Journal of Robotics Research
  23~(10-11) (2004) 1013--1027.

\bibitem{Kolt_Intelligent_control}
L.~P. III, Intelligent control and force redistribution for a high-speed
  quadruped trot, Ph.D. thesis, The Ohio State University (2007).

\bibitem{Kolt_Thrust_control}
J.~Estremera, K.~Waldron, Thrust control, stabilization and energetics of a
  quadruped running robot, The International Journal of Robotics Research
  27~(10) (2008) 1135--1151.

\bibitem{HyQ_Is_active_impedance}
C.~Semini, V.~Barasuol, T.~Boaventura, M.~Frigerio, J.~Buchli, Is active
  impedance the key to a breakthrough for legged robots?, in: Robotics
  Research, Springer, 2016, pp. 3--19.

\bibitem{StarlETH_Co_design_and_control}
M.~Hutter, Starleth, co-design and control of legged robots with compliant
  actuation, D (2013).

\bibitem{StarlETH_Control_of_dynamic_gaits}
C.~Gehring, S.~Coros, M.~Hutter, M.~Bloesch, M.~Hoepflinger, R.~Siegwart,
  Control of dynamic gaits for a quadrupedal robot, in: Robotics and automation
  (ICRA), 2013 IEEE international conference on, IEEE, 2013, pp. 3287--3292.

\bibitem{Cheetah_cub_Towards_dynamic_trot}
A.~Spröwitz, A.~Tuleu, M.~Vespignani, M.~Ajallooeian, E.~Badri, A.~Ijspeert,
  Towards dynamic trot gait locomotion: Design, control, and experiments with
  cheetah-cub, a compliant quadruped robot, The International Journal of
  Robotics Research 32~(8) (2013) 932--950.

\bibitem{BigDog}
R.~Playter, M.~Buehler, M.~Raibert, Bigdog, in: Unmanned Systems Technology
  VIII, Vol. 6230, International Society for Optics and Photonics, 2006, p.
  62302O.

\bibitem{Bigdog_the_rough_terrain}
M.~Raibert, K.~Blankespoor, G.~Nelson, R.~Playter, Bigdog, the rough-terrain
  quadruped robot, IFAC Proceedings Volumes 41~(2) (2008) 10822--10825.

\bibitem{Meet_Boston_dynamics_LS3}
K.~Michael, Meet boston dynamics' ls3-the latest robotic war machine.

\bibitem{Introducing_Spot}
BostonDynamics, Introducing spot,
  \url{https://www.youtube.com/watch?v=M8YjvHYbZ9w}, accessed February 9, 2015.

\bibitem{Billy_quadruped_robot}
Yobotics, Billy quadruped robot,
  \url{http://v.youku.com/v_show/id_XMzQ2NzkxMTEwNA==.html}, accessed March 16,
  2018.

\bibitem{Quadruped_Trotting_pacing_and}
M.~Raibert, Trotting, pacing and bounding by a quadruped robot, Journal of
  biomechanics 23 (1990) 7983--8198.

\bibitem{Quadruped_Running_on_four_legs}
M.~Raibert, M.~Chepponis, H.~Brown, Running on four legs as though they were
  one, IEEE Journal on Robotics and Automation 2~(2) (1986) 70--82.

\bibitem{HyQ_A_reactive_controller_framework}
V.~Barasuol, J.~Buchli, C.~Semini, M.~Frigerio, E.~D. Pieri, D.~Caldwell, A
  reactive controller framework for quadrupedal locomotion on challenging
  terrain, in: Robotics and Automation (ICRA), 2013 IEEE International
  Conference on, IEEE, 2013, pp. 2554--2561.

\bibitem{Scout_Some_finite_state_aspects}
R.~McGhee, Some finite state aspects of legged locomotion, Mathematical
  Biosciences 2~(1-2) (1968) 67--84.

\bibitem{Speed_stride_frequency_and}
N.~Heglund, C.~Taylor, Speed, stride frequency and energy cost per stride: how
  do they change with body size and gait?, Journal of Experimental Biology
  138~(1) (1988) 301--318.

\bibitem{Biped_walking_pattern_generation}
S.~Kajita, F.~Kanehiro, K.~Kaneko, K.~Fujiwara, K.~Harada, K.~Yokoi,
  H.~Hirukawa, Biped walking pattern generation by using preview control of
  zero-moment point, in: 2003 IEEE International Conference on Robotics and
  Automation (Cat. No. 03CH37422), Vol.~2, IEEE, 2003, pp. 1620--1626.

\end{thebibliography}

\end{document}